\documentclass[letterpaper, 10 pt, journal, twoside]{IEEEtran}

\usepackage{flushend}
\usepackage{graphicx}
\graphicspath{{images/}}
\usepackage{algorithm}
\usepackage{algorithmicx}
\usepackage{algpseudocode}
\usepackage{amsmath}
\usepackage{amssymb}
\usepackage{nicefrac}
\usepackage{amsfonts}
\usepackage{mathtools}
\usepackage{siunitx} %for
\usepackage{textcomp}%degree symbol
\usepackage{svg}
\usepackage{algpseudocode}

\usepackage{comment}

\usepackage{nomencl}
\makenomenclature
\usepackage{pifont}% http://ctan.org/pkg/pifont
\newcommand{\xmark}{\ding{55}}%

\usepackage{tablefootnote}
\usepackage{tabularx,booktabs}
\usepackage{gensymb}
\usepackage{multirow}
\usepackage[caption=false,font=footnotesize]{subfig}
\usepackage{arydshln}
\usepackage[pagebackref=false,breaklinks=true,colorlinks=true,bookmarks=false]{hyperref}
\usepackage{subfiles}
\usepackage{titling}  % if you need separate title pages for the subfiles
\date{}
\usepackage[noadjust]{cite}

\begin{document}

\bstctlcite{IEEEexample:BSTcontrol}

%%%%%%%%% TITLE
\title{\Huge Tracking Passengers and Baggage Items using \\ Multiple Overhead Cameras at Security Checkpoints}%
% Authors at the same institution
%\author{Abubakar Siddique and Henry Medeiros\\
%Marquette University, Milwaukee WI 53233 USA\\
%{\tt\small \{abubakar.siddique,henry.medeiros\}@marquette.edu}
%}
\author{Abubakar Siddique,~\IEEEmembership{Student Member,~IEEE,}
          Henry Medeiros,~\IEEEmembership{Senior Member, ~IEEE}% <-this % stops a space
%\thanks{Department of Electrical and Computer Engineering, Marquette University, Milwaukee, WI, 53233, e-mail: abubakar.siddique@marquette.edu \hspace{2cm} henry.medeiros@marquette.edu.}% <-this % stops a space
\thanks{Manuscript received August 22, 2022; accepted November 11, 2022. Date
of publication December 14, 2022.}
\thanks{This material is based upon work supported by the U.S. Department of Homeland Security, Science and Technology Directorate, Office of University Programs, under Award Number 2013-ST-061-E0001-04. The views and conclusions contained in this document are those of the authors and should not be interpreted as necessarily representing the official policies, either expressed or implied, of the U.S. Department of Homeland Security.}
%\thanks{This article was recommended by Associate Editor XX. YY. (Corresponding author: Abubakar Siddique)}
\thanks{Abubakar Siddique is with the Department of Electrical and Computer Engineering, Marquette University, Milwaukee, USA, e-mail: abubakar.siddique@marquette.edu}% 
\thanks{Henry Medeiros is with the Department of Agricultural and Biological Engineering, University of Florida, Gainesville, USA, e-mail: hmedeiros@ufl.edu} %
\thanks{Digital Object Identifier (DOI): 10.1109/TSMC.2022.3225252}
}
% The paper headers
\markboth{IEEE Transactions on Systems, Man, and Cybernetics: Systems,~Vol.~X, No.~Y, December 2022}{Siddique and Medeiros: Tracking Passengers and Baggage Items using Multiple Overhead Cameras at Security Checkpoints}
\maketitle

%%%%%%%%% ABSTRACT
\begin{abstract}
We introduce a novel framework to track multiple objects in overhead camera videos for airport checkpoint security scenarios where targets correspond to passengers and their baggage items. We propose a Self-Supervised Learning (SSL) technique to provide the model information about instance segmentation uncertainty from overhead images. Our SSL approach improves object detection by employing a test-time data augmentation and a regression-based, rotation-invariant pseudo-label refinement technique. Our pseudo-label generation method provides multiple geometrically-transformed images as inputs to a Convolutional Neural Network (CNN), regresses the augmented detections generated by the network to reduce localization errors, and then clusters them using the mean-shift algorithm. The self-supervised detector model is used in a single-camera tracking algorithm to generate temporal identifiers for the targets. Our method also incorporates a multi-view trajectory association mechanism to maintain consistent temporal identifiers as passengers travel across camera views. An evaluation of detection, tracking, and association performances on videos obtained from multiple overhead cameras in a realistic airport checkpoint environment demonstrates the effectiveness of the proposed approach. Our results show that self-supervision improves object detection accuracy by up to $42\%$ without increasing the inference time of the model. Our multi-camera association method achieves up to $89\%$ multi-object tracking accuracy with an average computation time of less than $15$ ms.
\end{abstract}
%%%%%%%%% BODY TEXT %%%%%%%%%%%%%%%%%%%%%%%%%%%%%%%%%%%%%%%%%%%%%
\section{Introduction}
\IEEEPARstart{A}{utomated} video surveillance requires the detection, tracking, and recognition of objects of interest in a scene. Accurate and precise surveillance in crowded scenes is one of the most challenging computer vision applications. To address the problem of visual surveillance in the domain of airport checkpoint security, the Department of Homeland Security (DHS) ALERT (Awareness and Localization of Explosives-Related Threats) center of excellence at Northeastern University initiated the CLASP (Correlating Luggage and Specific Passengers) project. This initiative aims to help the Transportation Security Administration (TSA) detect security incidents, such as theft of items and abandoned bags.

Current approaches for detecting and tracking passengers and luggage in airport checkpoints divide the image area within each camera's field of view into regions of interest where certain passenger behaviors are expected (e.g., passengers divest their items near the roller conveyor) \cite{5981718,islam2018corr}. While these approaches are effective within individual regions of interest,  they cannot detect and track passengers and their belongings throughout an entire checkpoint. Moreover, most recent detection algorithms \cite{MRCNN,DBLP:journals/corr/LiuAESR15,panet_liu2018,DBLP:journals/corr/LinMS016} are unable to detect multiple objects in realistic overhead camera scenarios due to the unavailability of large-scale datasets obtained using unconventional camera perspectives. 

Fine-tuning pre-trained models using human annotated labels is a common approach in computer vision methods. However, this strategy hinders the applicability of state-of-the-art algorithms in scenarios where images are obtained from perspectives that are not commonly observed in existing publicly available datasets.
%, and deployment-specific training data is needed.  
The dramatic variability of video surveillance systems used in airport checkpoints would require deployment-specific fine-tuning of models, and in some scenarios, even camera-specific adjustments. %This is currently the main barrier to the widespread adoption of models based on deep neural networks for such applications. 
To overcome this challenge, we leverage the fact that models pre-trained on large-scale datasets can build upon their initial predictions to adapt to new scenarios using SSL strategies. Our proposed SSL framework obviates the tedious and expensive human annotation procedure by automatically generating pseudo-labels to update the model. 
%while minimizing the instance uncertainty aware (cluster score iteratively guide the model to learn the equivariant transformation) multiple head losses.

%SSL Framework Figure: file size: 1.3 MB
\begin{figure*}[t]
\includegraphics[width=\textwidth]{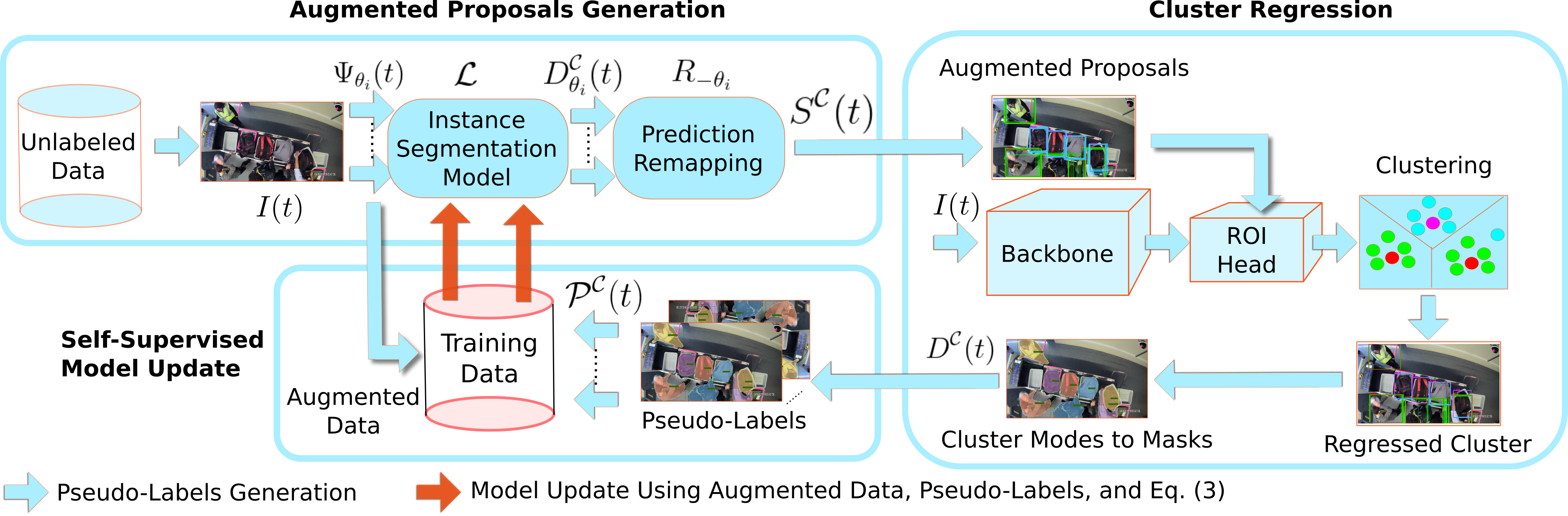}
\centering
\caption{Proposed SSL framework. The augmented proposal generation stage uses multiple rotated versions of the unlabeled input images to generate augmented detections from an instance segmentation model and then remaps these predictions into their original coordinates. The clustering algorithm leverages the model's regression ability to reduce localization errors using the augmented predictions as region proposals. The regressed cluster modes are then used to generate augmented pseudo-labels to update the model.}
\label{fig:model}
\end{figure*}

To generate pseudo-labels, we cluster multiple detections obtained from  geometrically transformed images using the mean-shift algorithm \cite{Mean-Shift-Comaniciu}. Each cluster corresponds to the detection of one object observed at different orientations on several augmented input images. The cluster modes with the corresponding bounding boxes, segmentation masks, and confidence scores are used to update the model. Thus, our model learns from rotation-invariant pseudo-labels and can be integrated with a tracking-by-detection algorithm \cite{tracktor_2019_ICCV} to generate accurate target tracklets from overhead perspectives.
%Closely integrated detector based tracker: Tracktor
%Detector is used for regressing tracks

Our SSL algorithm is inspired by the methods described in \cite{SSL_dets_2019_CVPR,DBLP:SSL_semantics,cai2021uncertainty_ssl, Wang_2021_ICCV_pseudo_labels_refine_semantics,mao2021noisy_annos}. However, unlike \cite{SSL_dets_2019_CVPR}, instead of resorting to multi-task strategies to guide the learning process, we employ a multi-inference approach similar in spirit to the self-consistency method based on equivariant transformations proposed in \cite{DBLP:SSL_semantics}. Our method differs from  \cite{DBLP:SSL_semantics} in that, rather than using the uncertainties from multiple model predictions to select image patches for additional training, it aggregates multiple inferences into accurate pseudo-labels that are used to refine the model.  Our method departs significantly from unsupervised model adaptation \cite{cai2021uncertainty_ssl} and knowledge distillation approaches \cite{Radosavovic2018Data} in that we only use automatically generated labels and avoid human annotations altogether during model update.  

We also propose a Multi-Camera Tracklet Association (MCTA) algorithm to maintain the temporal identifiers of passengers across cameras. We leverage the fact that our system is comprised of overhead cameras with partially overlapping fields of view to employ a simple but effective geometry-based trajectory association method. Our algorithm compares the projected centroids of target detections on neighboring cameras using the homographies between their image planes. We track passengers and bags across multiple views and generate global tracks by combining pairwise associations from the partially overlapping camera views. 

We evaluate the detection and tracking performance of our algorithms on videos from a simulated airport checkpoint and demonstrate that our approach performs on par with a model trained in an entirely supervised manner and substantially outperforms the pre-trained detection model. Our multi-camera evaluation shows that our MCTA method effectively handles the problem of passenger identity hand-off across cameras.

In summary, the key contributions of this work are:
\begin{itemize}
\item A novel self-supervised object detection algorithm that generates pseudo-labels based on instance segmentation uncertainties.
\item A new data augmentation and regression-based clustering mechanism that substantially improves the quality of pseudo-labels for self-supervised training.
%\item A robust and accurate multiple object tracking algorithm for overhead camera scenarios.
\item A new recursive tracklet association algorithm to address the identity hand-off issue during transitions between crowded overhead camera views.
%\item A distance-based association algorithm to keep track of the ownership of each baggage item across an airport checkpoint.
\item We provide an extensive evaluation of our methods on a dataset collected using multiple overhead cameras in a realistic airport checkpoint scenario.
\item Our SSL models and the corresponding source code are available at \url{https://github.com/siddiquemu/SCT_MCTA}.
\end{itemize}
%update this section after finishing the paper
To our knowledge, this is the first approach to solve the overhead multi-view association problem in a network of cameras with partially overlapping fields of view using a self-supervised detection strategy. 

\section{Related Work}
\label{sec:relatedwork}
Multiple target tracking using camera networks is an active research topic with several potential applications \cite{MAZZON201341, ZHANG201564, Hong2016Resource, Medeiros2008Distributed}. Most works on camera networks focus on the multi-camera aspect of the problem and do not consider the challenges associated with camera perspectives. Although generic object tracking algorithms could be used in surveillance systems (e.g. \cite{Wang_2019_CVPR,mozhdehi2018,JALILMOZHDEHI2022103479}), when object categories are known, trackers based on specialized detectors are more accurate and less prone to model drift \cite{fish_tracking_system,car_tracking_system}. This observation has led to the development of a variety of multiple target tracking algorithms that specialize in tracking humans \cite{henschel2019multiple,sadeghian2019sophie,alahi2016social,gupta2018social,babaee2018multiple,motion_seg_cor_cocluster,MV_track_3D_homog,wildtrack_multicam,MV_deepocclusion_3D, subspace_learning_person_reid2020, statistical_feature_learning_personReID2018} or vehicles \cite{MV_Reid_cam_model,MV_reid_feature,MV_ssl_reid,vehicle_visual_tracking2021}. However, in many scenarios, it is desirable to track additional objects of known categories. In these cases, more flexible detection algorithms are needed, but the effectiveness of modern object detection models is highly dependent upon the characteristics of the training datasets \cite{DBLP:journals/corr/LiuAESR15,MRCNN,panet_liu2018}.

Previous works have used SSL techniques to improve visual feature learning 
%by solving some predefined auxiliary tasks
\cite{SSl_survey_TPAMI2020, SSL_backbone_deep_cluster, DBLP:SSL_backbone_wild}, reducing dependency on human annotations for training backbone models. %Self-supervision techniques for 
However, transferring knowledge from pre-trained backbones to downstream tasks is a far less explored topic. Unlike our proposed approach, SSL techniques for detection \cite{SSL_dets_2019_CVPR, cai2021uncertainty_ssl} and semantic segmentation \cite{DBLP:SSL_semantics} rely on annotations to initialize the model before iterative learning can take place. 

\begin{figure*}[t]
\centering
%file size 1.2 MB
\includegraphics[width=0.95\textwidth]{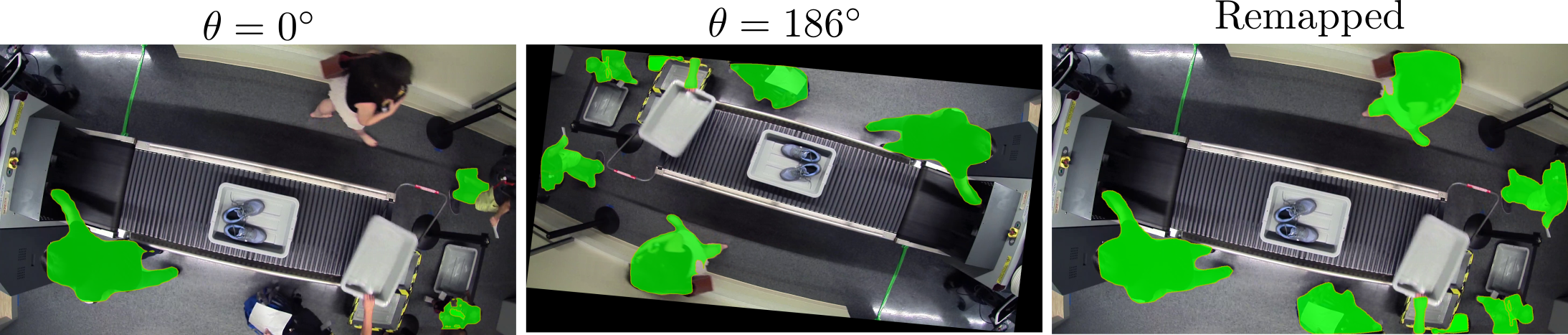} \\
\vspace{0.07cm}  
\includegraphics[width=.95\textwidth]{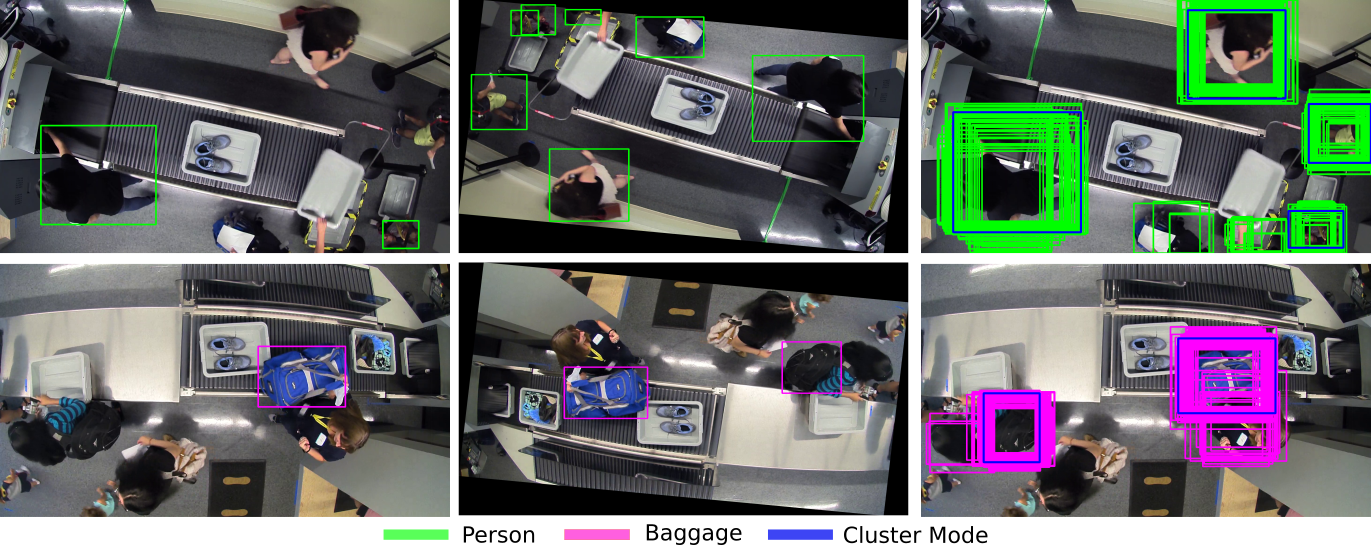}
%file size: 1 MB
\caption {Visualization of our data augmentation approach. The first and second columns show the segmentation masks and detections 
%(green for person, magenta for baggage)
at $\theta = 0^{\circ}$ and $\theta = 186^{\circ}$, respectively. The third column shows the remapped detections in the set $S^{\mathcal{C}}$ on the original image (using Alg. \ref{alg:augmentedproposals}) with the best detections (blue) from Alg. \ref{alg:clusterregression}.} 
\label{fig:det_model_describe}
\end{figure*} 

Data augmentation is an effective mechanism to improve the robustness of CNNs in scenarios not available during training \cite{Radosavovic2018Data, LIU2016Upright}, but little attention has been given so far to approaches for combining the response of the network to augmented samples. 
%The performance of such approaches depends on the mechanisms used to combine the response of the network to the augmented samples. 
In multi-target tracking applications, multiple detections mapped to a common coordinate system can be interpreted as the probability of occupancy of the area observed by the cameras \cite{multicamera-Mean-Shift-Taj}. Although it is possible to use clustering techniques to map the modes of this distribution to unique target detections, bounding box alignment errors pose a challenge to the generation of high-quality pseudo-labels for SSL. Hence, we propose a test-time regression technique that leverages instance segmentation information for pseudo-label generation.

A systematic solution to the data association problem is another important component of multi-target tracking-by-detection methods \cite{dual_attention,quad_CNN,sun2019deep,sheng2018iterative, sheng2018heterogeneous,kim2018multi,fang2018recurrent,schulter2017deep,MOT_spatio-temporal_embed,MOTS,Kim:2015:MHT:2919332.2920100,track_untrackable,ding2016new, siddiqueBMVC2021_usc_mots}. Single-camera trackers \cite{tracktor_2019_ICCV, Stadler_2021_CVPR_tracktor} use detectors trained on multiple datasets \cite{DBLP:journals/corr/Leal-TaixeMRRS15} to generate bounding boxes and form track hypotheses for all the targets in each frame. In this work, we employ a state-of-the-art single-camera tracker \cite{tracktor_2019_ICCV} using a detector based on our self-supervised models, which achieves unprecedented tracking performance in previously unseen airport surveillance videos.

Finally, multi-camera tracking systems require sophisticated trajectory association mechanisms to maintain target identities across cameras \cite{MCMT_hierarchy,Tracklet_Cavallaro,Nithin17Multi}. Even within a single camera, occlusions must be handled using similar strategies \cite{tracklet_association,tracklet_asso_hausdorff}. Most association approaches compute trajectory similarity scores based on a combination of appearance and motion features \cite{MCMT_hierarchy,Tracklet_Cavallaro,Nithin17Multi,tracklet_association,tracklet_asso_hausdorff}. These features are learned using a large number of continuous trajectories, which are difficult to obtain with typical ceiling-height overhead cameras due to their limited fields of view. %Multi-camera methods also need to project trajectories onto a global 3D coordinate frame, which requires camera calibration information. 
Some methods use camera calibration information to project tracks onto a common plane and perform association using occlusion modeling  \cite{MV_deepocclusion_3D} or re-identification techniques \cite{MV_reid_feature,statistical_feature_learning_personReID2018,MV_ssl_reid, MV_Reid_cam_model,subspace_learning_person_reid2020}. Dependency on camera calibration further limits the applicability of these methods to security systems since calibrating multiple cameras with partially overlapping fields of view is a complex task \cite{neural_reproj_2021_CVPR, deep_homography_2020_eccv,bok2014extrinsic,medeiros2008online}. 
%%%%%%%%%%%%%%%%%%%%%%%%%%%%%%%%%%%%%%%%%%%%

\section{Proposed Model}
\label{sec:proposedmodel}
Our system consists of two main components: i) a detection algorithm trained using SSL and ii) a multi-camera tracking-by-detection mechanism. A single-camera tracking algorithm uses SSL detections to generate tracklets for passengers and baggage items. We then employ a novel multi-camera target trajectory association algorithm to uniquely identify passengers throughout the checkpoint. 
%-------------------------------------------------------------------------
\subsection{Self-Supervised Learning}
We use the PANet model \cite{panet_liu2018} with a ResNet-50 backbone \mbox{\cite{ResNeXt,DBLP:journals/corr/HeZRS15}} as the baseline detector. Since the categories of interest are persons and their belongings, we use a model pre-trained on the COCO dataset \cite{coco_data}, which includes object classes related to these categories (i.e., person, handbag, backpack, and suitcase). Because the COCO dataset consists mostly of images captured at roughly eye-level, detectors trained using that dataset do not perform well on overhead perspectives.
To address this limitation, our SSL framework updates the baseline model using rotation-invariant pseudo-labels. As Fig. \ref{fig:model} shows, our SSL framework consists of three main steps: i) augmented region proposals generation, ii) pseudo-label generation and refinement through cluster regression, and iii) iterative model update. 

%Regression example
\begin{figure}[b]
\centering
%file size 663 KB
\includegraphics[trim=0 0 0 0cm,clip,width=.95\linewidth]{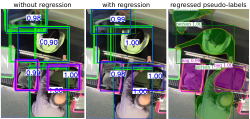}
\caption {Regression on test-time augmented bounding boxes (middle) and cluster modes (right) to generate pseudo-labels for SSL training.}
\label{fig:regression_example}
\end{figure} 
%TODO: Do we need to write algorithm including SSL iterations as well!
\begin{algorithm}[ht]
\caption{Augmented Proposals Generation} \label{alg:augmentedproposals}
     \begin{algorithmic}[1]
         \Function{AugmentedProposals}{$I(t)$, $r$}
         %\State {$D_p(t) = D_{MRCNN}(I(t))$}
            \State {$S^{\mathcal{C}}(t)=\emptyset$, $\Theta = \left\{i\cdot\Delta\theta\right\}_{i=1}^r$}
            %\State {$\Phi(t) = \text{ROI}_{r}(I(t))$} 
            \For {$\theta_i\in \Theta$}                      
              \State {$\Psi_{\theta_i}(t) = R_{\theta_i} (I(t))$}
              \State {$D^{\mathcal{C}}_{\theta_i}(t) = D_{\text{PANet}}(\Psi_{\theta_i}(t))$}
              \State {$S^{\mathcal{C}}_{\theta_i}(t) = R_{-\theta_i}(D^{\mathcal{C}}_{\theta_i}(t))$}
              \State {$S^{C}(t) = S^{C}(t) \cup S^{\mathcal{C}}_{\theta_i}(t)$} 
              %\State {$\theta_i= \theta_{i-1}+\Delta\theta$}            
            \EndFor
            \State \Return $S^{C}(t)$
        \EndFunction
   \end{algorithmic}
\end{algorithm}

\subsubsection{Augmented Proposals Generation}
Our data augmentation method, summarized in Alg. \ref{alg:augmentedproposals}, uses the PANet model to detect and segment multiple instances of objects of interest. During the first iteration of SSL training, we retain only the outputs of the pre-trained model for the \emph{person}, \emph{handbag}, \emph{backpack}, and \emph{suitcase} classes. The \emph{person} class corresponds to passengers and detections of \emph{handbag}, \emph{backpack}, and  \emph{suitcase} items are treated as baggage items. In subsequent iterations of SSL training, we modify the model to generate only the object categories \nomenclature{$\mathcal{C}$}{object category $\mathcal{C} \in \{pax,bag\}$ (i.e., person or baggage)} $\mathcal{C} \in \{pax,bag\}$, where $pax$ corresponds to passengers and $bag$ to baggage items. Let \nomenclature{$D^{\mathcal{C}}(t)$}{Set of detections on image $I(t)$ at time $t$ of object class $\mathcal{C}$} $D^{\mathcal{C}}(t)$ be the set of detections on image $I(t)$ at time $t$. That is, $D^{\mathcal{C}}(t) = \{d_1, \ldots, d_{n_t^\mathcal{C}}\}$, where \nomenclature{$d_j$}{Detection of the $j$-th object and $n_t^\mathcal{C}$ is the number of objects of class $\mathcal{C}$ in frame $I(t)$} $d_j \in \mathbb{R}^5$ is the detection of the $j$-th object and $n_t^\mathcal{C}$ is the number of objects of class $\mathcal{C}$ in frame $I(t)$. Each detection $d_j$ consists of the coordinates and dimensions of the target's bounding box, \nomenclature{$b_j^\mathcal{C}$}{4D vector comprising the coordinates and dimensions of the $j$-th target's bounding box} $b_j^\mathcal{C}\in \mathbb{R}^4$, as well as its detection confidence score \nomenclature{$s_j$}{Confidence score of j-th detection $d_j$} $s_j\in [0,1]$.

We noticed that the detector performs better when objects are observed at more commonly occurring angles (e.g., upright). Therefore, to reduce the negative effect of the overhead perspective, we generate \nomenclature{$\Psi_{\theta_i}(t)$}{Copy of the image $I(t)$ rotated by an angle $\theta_i$} multiple rotated copies of the input image $\Psi_{\theta_i}(t)=R_{\theta_i} (I(t))$ (line 4 in Alg. \ref{alg:augmentedproposals}), where \nomenclature{$R_{\theta_i}(\cdot)$}{Rotation operator to rotate the image by an angle $\theta_i$} $R_{\theta_i}(\cdot)$ is the rotation operator, which rotates the image by an angle $\theta_i$. The angle of rotation $\theta_i$ varies between $0$ and $2\pi$ at intervals of \nomenclature{$\Delta\theta$}{Rotation step in augmented proposals generation} $\Delta\theta =\left \lfloor \frac{2\pi }{r} \right \rfloor$, \nomenclature{$r$}{Number rotation angles used in the augmentation} i.e., $\theta_i = \Delta\theta, \ldots, 2\pi$, where  $r$ determines the rotation resolution. At each rotation step, we compute the \nomenclature{$D^{\theta_i}_{\mathcal{C}}(t)$}{Detection set  for classes $\mathcal{C}\in\{pax,bag\}$ in the rotated ROI $\Psi_{\theta_i}(t)$} detection set $D_{\theta_i}^{\mathcal{C}}(t)$ for both classes $\mathcal{C}\in\{pax,bag\}$ using a single call to the function \nomenclature{$D_{\text{PANet}}(\cdot)$}{PANet as a baseline detector} $D_{\text{PANet}}(\cdot)$ (line 5). We then remap the resulting detections to the coordinate frame of the original image by applying the inverse rotation to each of the detections in $D_{\theta_i}^{\mathcal{C}}(t)$ (line 6). To avoid localization errors introduced by rotating axis-aligned bounding boxes, we apply the rotation operation to the binary segmentation masks produced by PANet and compute the corresponding bounding boxes using the rotated masks. At the end of Alg. \ref{alg:augmentedproposals}, the set \nomenclature{$S^{C}(t)$}{Se of all augmented detections from all the rotation angles $\theta_i$}  $S^{C}(t) = \cup_{i=1}^{r} S^{\mathcal{C}}_{\theta_i}(t)$
contains the detections at all the rotation angles $\theta_i$. Fig. \ref{fig:det_model_describe} illustrates the detections at two rotation angles and the result of mapping detections at 20 different orientations back to the original coordinate system.

\begin{figure}[t]
\centering
\subfloat{\includegraphics[width=.95\linewidth]{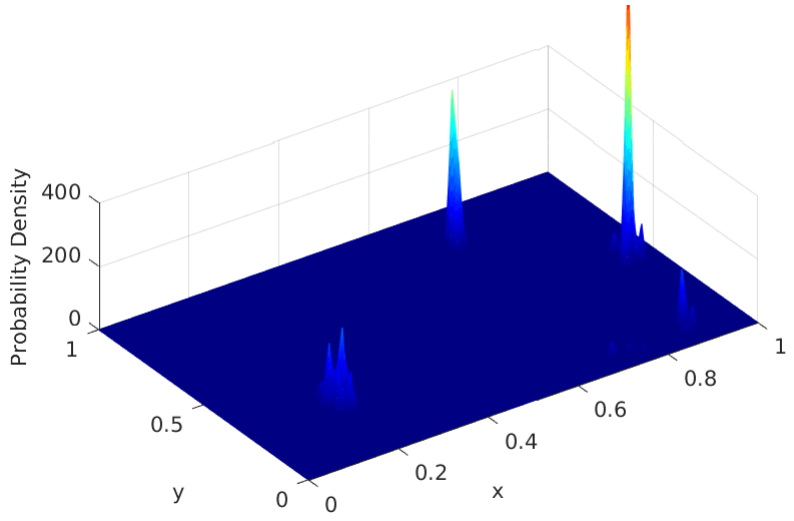}}
%\subfloat{\includesvg[width=.48\linewidth]{svgfiles/fr_4626_all.svg}}
\caption {Probability of occupancy of passengers at one frame of our evaluation datasets (Fig. \ref{fig:det_model_describe}).}
%\caption {Probability of occupancy of passengers (left) and baggage (right) at one frame of our evaluation datasets (Fig. \ref{fig:det_model_describe}).}
% using remapped set $S_{\mathcal{C}}$ in dataset A.
\label{fig:distribution}
\end{figure} 

%TODO: Do we need to write algorithm including SSL iterations as well!
\begin{algorithm}[ht]
\caption{Cluster Regression} \label{alg:clusterregression}
     \begin{algorithmic}[1]
         \Function{ClusterRegression}{$S^{C}(t)$}
         %\Require{Set of augmented detections $S^{C}(t)$}
         %\Ensure{Set of regressed detections $\mathcal{D}^{C}(t)$}
         %\State {$D_p(t) = D_{MRCNN}(I(t))$}
           \State $\mathcal{D}^\mathcal{C}(t)=\emptyset$
           \State \parbox[t]{\dimexpr\linewidth-\algorithmicindent}{Refine the augmented detections using $S^{\mathcal{C}}(t)$ as region proposals for the $D_{\text{PANet}}$ model}
           \State {$O^{\mathcal{C}}(t) = \mathtt{mean-shift}(S^{\mathcal{C}}(t))$}
           \For {$Q \in O^\mathcal{C}(t)$}
			\State {Compute the cluster score $\bar{\eta}_Q$ using Eq. \ref{eq:cluster_score}}
            %\underline{\State {$\bar{\eta}=\frac{\sum_{d_i\in Q} \mathtt{score}(d_i)}{ \mathtt{k}}$}}
            \If {$\bar{\eta}_Q \geq \lambda$}
             \State {$d = \underset{d_i\in Q} {\mathrm{argmax}} \left (s_i \right )$}
             \State {$\mathcal{D}^\mathcal{C}(t) = \mathcal{D}^\mathcal{C}(t) \cup \{d\}$} 
            \EndIf
           \EndFor
           \State \Return $\mathcal{D}^{C}(t)$
           \EndFunction
   \end{algorithmic}
\end{algorithm}

\subsubsection{Cluster Regression}
\label{sec:cluster-regression}
 %sIMILAR DISCUSSIONS IN CLUSTER REGRESSION SECTION??
 %Although the remapping of the augmented detections using segmentation mask is applicable for clustering, it still shows (Fig. \ref{fig:det_model_describe}) localization error that we need to correct for using cluster modes as pseudo-labels in SSL. 
Alg. \ref{alg:clusterregression} summarizes our approach to combine the set of augmented detections  $S^{C}(t)$ into a set of refined target detections  $\mathcal{D}^{C}(t)$. To reduce discrepancies among bounding boxes caused by segmentation errors, we leverage the pre-trained model to regress the set of augmented detections $S^{C}(t)$. As shown in Fig. \ref{fig:model}, our cluster regression method uses the backbone features \cite{FPN}  with the augmented detections $S^{C}(t)$ as region proposals (instead of proposals generated using the region proposal network \cite{FRCNN}) to the downstream box and mask heads. To avoid disregarding low-confidence detections that might correspond to relevant region proposals, we do not apply non-maximum suppression to the model predictions.
Fig. \ref{fig:regression_example} shows that cluster regression significantly increases the accuracy of the bounding boxes generated using the augmented input, and the corresponding segmentation masks are consequently also more accurate.%  We use the regressed augmented detection scores in the clustering and then compute the cluster score to measure the rotational invariance (see section \ref{sec:modified_loss}). 

\paragraph{Cluster Mode Detection}
As Fig. \ref{fig:distribution} illustrates,  detections and their corresponding confidence scores form a non-parametric distribution of the image's occupancy probability. We use the mean-shift algorithm \cite{multicamera-Mean-Shift-Taj} to identify the modes of that distribution and cluster detections corresponding to common targets. We cluster detections according to their bounding boxes $b_j$ using a multivariate Gaussian kernel \cite{multicamera-Mean-Shift-Taj} with bandwidth \nomenclature{$h^\mathcal{C}$}{Bandwidth of multivariate Gaussian kernel} $h^\mathcal{C}$. We use the sample variances of the object bounding boxes at each frame to determine the kernel bandwidth, i.e., 
\begin{equation}
h^\mathcal{C} =\text{diag} \left( \sum_{j=1}^{n^\mathcal{C}_t}{(b^{\mathcal{C}}_j - \bar{b}^{\mathcal{C}}_j)(b^{\mathcal{C}}_j - \bar{b}^{\mathcal{C}}_j)^T}\right),
\label{eq:sigmac}
\end{equation}
where $\bar{b}^{\mathcal{C}}_j$ is the sample mean of $b^{\mathcal{C}}_j$ and $\text{diag} \left(\cdot \right)$ is the diagonal of the covariance matrix. The correlations among the elements of $b_j$ are negligible and can be safely ignored.
%NOTE: Why don't we simply compute the sample variance (b_i - \bar{b_i})(b_i - \bar{b_i})^T?
Each call to the mean-shift algorithm (line 4 in Alg. \ref{alg:clusterregression}) produces a set of clusters \nomenclature{$O^\mathcal{C}(t)$}{Set of clusters from mean-shift algorithm for each class $\mathcal{C}$ whose elements are sets of detections assigned to the same target} $O^\mathcal{C}(t)$ whose elements are sets of detections assigned to the same target. We consider the detections of passengers and baggage items separately. Hence, two separate invocations of the mean-shift procedure are required to produce the sets \nomenclature{$O^{pax}(t)$}{separate set of passengers} $O^{pax}(t)$ and \nomenclature{$O^{bag}(t)$}{separate set of baggage} $O^{bag}(t)$. The confidence score \nomenclature{$\bar{\eta}_Q$}{Confidence score of cluster $Q\in O^{\mathcal{C}}(t)$} $\bar{\eta}_Q$ of cluster $Q\in O^{\mathcal{C}}(t)$ is defined as the ratio between the total score of detections within that cluster and the number of rotation angles considered in the augmentation process, i.e.,
\begin{equation}
\bar{\eta}_Q = \frac{1}{r}\ensuremath{\sum_{d_{j}\in Q}s_{j}}.
\label{eq:cluster_score}
\end{equation}
Lines 6-10 of Alg. \ref{alg:clusterregression} show that we discard clusters with scores lower than a threshold $\lambda$ to remove false positive detections.

%TODO: Do we need to write algorithm including SSL iterations as well!
\begin{algorithm}[t]
\caption{Pseudo-Label Generation} \label{alg:pseudolabel}
     \begin{algorithmic}[1]
         \Function{PseudoLabels}{$\mathcal{D}^{C}(t)$, $r$}
            \State $\mathcal{P}^\mathcal{C}(t)=\emptyset$,  $\Theta = \left\{i\cdot\Delta\theta\right\}_{i=1}^r$
            \For {${d_j}\in D^\mathcal{C}(t)$}        
                \For {$\theta_i\in \Theta$}        
                    \State \parbox[t]{\dimexpr\linewidth-\algorithmicindent}{Generate the augmented region proposals \\
                $d_{i,j}=R_{\theta_i} (d_j)$}
                \EndFor
                \State \parbox[t]{\dimexpr\linewidth-\algorithmicindent}{Generate the pseudo-label $(\hat{b}_{i}, \hat{m}_i, \alpha_i)$ using the \\ region proposals $d_{i,j}$}
                \State {$\mathcal{P}^{C}(t) = \mathcal{P}^{C}(t) \cup \left\{(\hat{b}_{i}, \hat{m}_i, \alpha_i)\right\}$} 
            \EndFor
            \State \Return $\mathcal{P}^{C}(t)$
            \EndFunction
   \end{algorithmic}
\end{algorithm}

\subsubsection{Self-Supervised Model Update}
Alg. \ref{alg:pseudolabel} shows the procedure to generate the pseudo-labels used to update the model. Since our goal is to train the model using labels generated from multiple perspectives, we rotate both the original image and the corresponding predicted modes to generate pseudo-label proposals at each orientation. That is, for each mode $d_j\in D^{\mathcal{C}}(t)$, we generate the pseudo-label mask \nomenclature{$\hat{m}_{j}$}{pseudo-label mask} $\hat{m}_{j}$ by using the rotated cluster modes $d_{i,j}=R_{\theta_i} (d_j)$, $i=1,\ldots,r$ as region proposals for the segmentation head, using the same approach described in Section \ref{sec:cluster-regression}. We then find the bounding box \nomenclature{$\hat{b}_{j}$}{Pseudo label bounding box} $\hat{b}_{j}$ corresponding to $\hat{m}_{j}$. \nomenclature{$\hat{\alpha}_j$}{Confidence score of the $j$-th pseudo-label} The confidence $\hat{\alpha}_j$ of the resulting pseudo-label is given by its corresponding cluster score. \nomenclature{$\mathcal{P}^{\mathcal{C}}(t)$}{Set of pseudo-labels for category $\mathcal{C}$ at time $t$} The set of pseudo-labels $\mathcal{P}^{\mathcal{C}}(t) = \left\{(\hat{b}_{j}, \hat{m}_j, \hat{\alpha}_j)\mid d_j\in  D^{\mathcal{C}}(t)\right\}$ thus contains accurate annotations even for targets that the model is unable to detect at certain orientations. 

\paragraph{Rotation-Invariant Loss}
\label{sec:modified_loss}

To update the model using  rotation-invariant pseudo-labels in a robust and efficient manner, we propose a novel uncertainty-aware, multi-task loss function given by
\begin{multline}
\mathcal{L}=\sum_{\hat{c}\in\mathcal{C}}\sum_{(\hat{b}_j,\hat{m}_j,\hat{\alpha}_j)\in\mathcal{P}^{\mathcal{C}}(t)}\hat{\alpha}_j(\mathcal{L}^{c}(\hat{c},\tilde{c})+\mathcal{L}^{b}(\hat{b}_j,\tilde{b}_j)\\
+\mathcal{L}^{m}(\hat{m}_j,\tilde{m}_j))  + \mathcal{L}_{rpn},
\label{eq:modified_panet_loss}
\end{multline}
%\begin{equation}
%\begin{aligned}
%\mathcal{L}_{total} &= \sum^{N}_{i=1} \alpha_i \left ( \mathcal{L}_i^c   + \mathcal{L}_i^b + \mathcal{L}_i^m \right) + \mathcal{L}_{rpn},
%\label{eq:modified_panet_loss}
%\end{aligned}
%\end{equation}
where \nomenclature{$\hat{c}$}{Object class for the pseudo-label} \nomenclature{$\tilde{c}$}{Object class predicted by the network} $\tilde{c}$, \nomenclature{$\tilde{b}_j$}{Object bounding box predicted by the network} $\tilde{b}_j$, and \nomenclature{$\tilde{m}_j$}{Object mask predicted by the network} $\tilde{m}_j$ are the object class, bounding box, and segmentation mask predicted by the network; \nomenclature{$\mathcal{L}^c$}{Classification loss} $\mathcal{L}^c$, \nomenclature{$\mathcal{L}^b$}{Box regression loss} $\mathcal{L}^b$,  and \nomenclature{$\mathcal{L}^m$}{Pixel-wise binary cross entropy loss} $\mathcal{L}^m$ are the  classification and bounding box regression losses defined in \cite{FRCNN} and the pixel-wise binary cross entropy mask loss described in \cite{MRCNN}; and $\mathcal{L}_{rpn}$ is the region proposal network loss from \cite{FRCNN}. In Eq. \ref{eq:modified_panet_loss},  the instance head losses are weighted by their corresponding cluster scores. This strategy ensures that instances with low cluster scores that might correspond to incorrect pseudo-labels have little impact on the update of the network parameters. %Thus, the network vanishes the uncertainty in the training data by learning the rotation invariant property.
As Alg. \ref{alg:detection} indicates, a new set of pseudo-labels is generated at each SSL iteration using the updated model from the previous iteration. 

%TODO: Do we need to write algorithm including SSL iterations as well!
\begin{algorithm}[ht]
\caption{Self-Supervised Detection Model Update} \label{alg:detection}
     \begin{algorithmic}[1]
         \Require{Image sequence $I(t)$, $t=1,\ldots,T$}
         \Ensure{Updated detection model $D_{\text{PANet}}$}
         %\State {$D_p(t) = D_{MRCNN}(I(t))$}
         \Repeat
             \For{ $t=1, \ldots, T$}
                \State $S^{\mathcal{C}}(t) =$ \Call{AugmentedProposals}{$I(t)$}
                \State $\mathcal{D}^{\mathcal{C}}(t) =$ \Call{ClusterRegression}{$S^{\mathcal{C}}(t)$}
                \State $\mathcal{P}^{\mathcal{C}}(t) =$ \Call{PseudoLabels}{$\mathcal{D}^{\mathcal{C}}(t)$}
            \EndFor
            \State  \parbox[t]{\dimexpr\linewidth-\algorithmicindent}{Fine-tune the $D_{\text{PANet}}$ model using the pseudo-labels $\left\{\mathcal{P}^{\mathcal{C}}(t)\right\}_{t=1}^T$ according to the loss function in Eq. \ref{eq:modified_panet_loss}}
        \Until Convergence criterion is met
   \end{algorithmic}
\end{algorithm}

\subsection{Multi-View Passenger and Baggage Tracking}
%TBD: Brief intro to the tracking algorithms.
Our multi-camera tracking framework comprises two main steps: i) a single-camera, multiple-target tracking-by-detection algorithm, and ii) a multi-camera trajectory association mechanism. Our single-camera tracker uses the detections generated by our SSL framework and a Single-Camera Trajectory Association (SCA) method to keep track of the identities of individual passengers and baggage items within the field of view of each camera. Our MCTA strategy then projects the trajectories of passengers observed in cameras with overlapping fields of view onto a common image plane. These trajectories are then compared using the Fr\'{e}chet distance and associated using a recursive graph-based mechanism.
%%%%%%%%%%%%% Tracking Algorithm %%%%%%%%%%%%%%%%%%%%%%%%%%%%%
%We need to describe the followings for SCT either in paper or in supple
\subsubsection{Single-camera Tracking}
\label{sec:sca}
We use the Tracktor algorithm \cite{tracktor_2019_ICCV} as our baseline single-camera tracker. The output of the algorithm at each image frame is a set %\nomenclature{$T$}{Final timestamp for a sequence} 
\nomenclature{$\mathcal{T}^{\mathcal{C}}(t)$}{Set of bounding boxes ${b}_j$ and their corresponding temporal identifier labels $l_j$} $T^{\mathcal{C}}(t) = \{ \omega_1, \ldots, \omega_{n_t^\mathcal{C}} \}$, where \nomenclature{$\omega_j$}{Passenger and baggage tracker bounding box and temporal identifier, $[{b}_j, l_j]$} $\omega_j = [{b}_j, l_j]$, \nomenclature{$l_j$}{unique identifier label for each passenger and baggage item in the frame} with $l_j$ corresponding to a unique identifier label for each passenger and baggage item in the frame. These labels remain the same throughout the video sequence and hence perform temporal association among detections. The tracklet for the $k$-th object is thus given by the set of detections over the entire video sequence whose temporal identifier is $l_j=k$, i.e., $\tau_{k}=\cup_{t=1}^{T}\left\{\omega_j\mid \omega_j\in T^{\mathcal{C}}(t), l_{j}=k\right\}$.
%NOTE: Since the tracktor already has an online  IoU and appearance-based monocular ReID, our SCTA will do the associations after generating the tracklets. We need to clarify the significance of SCTA technique.
Temporary occlusions between passengers may lead to the fragmentation of trajectories within the field of view of a camera. Tracktor's  simple re-identification strategy is unable to accommodate the longer occlusions, appearance variations, and somewhat erratic motion patterns commonly observed in airport checkpoints. Thus, we incorporate an SCA \nomenclature{SCA}{single-camera trajectory association algorithm} mechanism to resolve this issue. Our method associates new tracklets with recently terminated tracklets such that the Euclidean distance 
%intersection over union (IoU) 
between the centroids of the last detection of the previous tracklet and the first detection of the new tracklet is minimized. That is, \nomenclature{$\mathcal{C}_{sc}(\tau_m,\tau_n)$}{Association cost of the distinct tracklets $\tau_m$ and $\tau_n$ in a single view} let $\tau_m$ and $\tau_n$ be two distinct tracklets, and \nomenclature{$b_m^{t^i}$}{First detection of tracklet $\tau_m$} $b_m^{t^i}$,  \nomenclature{$b_n^{t^f}$}{last detection of $\tau_n$} $b_n^{t^f}$ be the first detection of $\tau_m$ and the last detection of $\tau_n$, respectively.  \nomenclature{$\delta_{e}$}{Euclidean distance between the centroids of the last detection of the previous tracklet, $b_m^{t^i}$ and the first detection of the new tracklet, $b_n^{t^f}$} Defining $\delta_{e}=||b_m^{t^i}-b_n^{t^f}||^2$, the association cost between $\tau_m$ and $\tau_n$ is given by
\begin{equation}
\mathcal{C}_{sc}(\tau_m,\tau_n)=\begin{cases} \delta_{e} & \text{ if } 0< t^i - t^f\leqslant t_{th} \text{, } \delta_{e} < \delta_{max} \\ \infty& \text{ otherwise,} \end{cases}
\label{eq:single_tracklet}
\end{equation}
where \nomenclature{$t_{th}$}{Maximum temporal offset to consider two tracklets for association} $t_{th}$, \nomenclature{$\delta_{max}$}{Maximum Euclidean distance to consider two tracklets for association} $\delta_{max}$ are the maximum temporal offset and maximum distance to consider two tracklets for association. We then compute the optimal tracklet assignment using the Hungarian algorithm based on the costs $\mathcal{C}_{sc}(\tau_m,\tau_n)$.

\subsubsection{Multi-Camera Tracklet Association}
%% MCMT algorithm
\begin{algorithm}
\caption{Multi-Camera Tracklet Association Algorithm} \label{alg:tracklet}
     \begin{algorithmic}[1]
         \Require{Set of tracklets from the primary camera $\mathcal{T}_p$ and the auxiliary camera $\mathcal{T}_a$, homography $H_{p,a}$ mapping the auxiliary camera image plane to that of the primary camera}
         \Ensure{Updated set of primary tracklet labels}
          \State \parbox[t]{\dimexpr\linewidth-\algorithmicindent}{Project the detections of tracklets in $\mathcal{T}_a$ onto the image plane of the primary camera using $H_{p,a}$}
          \State \parbox[t]{\dimexpr\linewidth-\algorithmicindent}{Compute the association costs $\mathcal{C}_{mc}(\tau_a, \tau_p)$ $\forall \tau_p \in \mathcal{T}_p$, $\forall \tau_a \in \mathcal{T}_a$ according to Eq. \ref{eq:cmc}}
        \State \parbox[t]{\dimexpr\linewidth-\algorithmicindent}{Initialize the graph $\mathcal{G}_{mc}=(V, E)$, $E=\emptyset$, $V=\{\tau| \tau\in \mathcal{T}_p \cup \mathcal{T}_a\}$}
        \While {$\min_{\tau_p \in \mathcal{T}_p, \tau_a \in \mathcal{T}_a}{\left(\mathcal{C}_{mc}(\tau_a, \tau_p) \right)} < \infty $}
        \State \parbox[t]{\dimexpr\linewidth-\algorithmicindent}{Associate tracklet segments using the Hungarian algorithm based on the costs $\mathcal{C}_{mc}$}
        \State \parbox[t]{\dimexpr\linewidth-\algorithmicindent}{Update the costs of the tracklets $\tau \in \mathcal{T}_a$ and $\tau' \in \mathcal{T}_p$ for which $\tau \cap \tau_a \notin \emptyset$ and $\tau' \cap \tau_p \notin \emptyset$ to $\mathcal{C}_{mc}(\tau,\tau_p)=\mathcal{C}_{mc}(\tau_a,\tau')=\infty$}
        \State \parbox[t]{\dimexpr\linewidth-\algorithmicindent}{$E = E \cup (\tau_a,\tau_p)$}
        \EndWhile
        \For {each $\tau_p \in \mathcal{T}_p$}
        \State \parbox[t]{\dimexpr\linewidth-\algorithmicindent}{$\mathcal{N}_{p} = \text{DFS}(\tau_p,\mathcal{G}_{mc})$}
        \State \parbox[t]{\dimexpr\linewidth-\algorithmicindent}{Update the labels of tracklets in $\mathcal{N}_{p}$ using Eq. \ref{eq:l_tau}}
        \State \parbox[t]{\dimexpr\linewidth-\algorithmicindent}{$E = E - \{(\tau_i,\tau_j) | (\tau_i,\tau_j) \in \mathcal{N}_{p}\}$}
        \EndFor
     \end{algorithmic}
\end{algorithm}

Since passengers may temporarily leave and later re-enter the fields of view of individual cameras, their corresponding trajectories may be fragmented into multiple segments. 
To associate tracklets across camera views, we consider the fact that two tracklets corresponding to the same target include temporally overlapping detections. Let the camera whose partial tracklets we wish to complete be our \emph{primary} camera, and let the \emph{auxiliary} camera be the one whose \nomenclature{$\mathcal{T}_p$}{Set of tracklets from the primary camera} tracklets will be used to complement the \nomenclature{$\mathcal{T}_a$}{Set of tracklets from the auxiliary camera} tracklets observed by the primary camera. Further, let $\mathcal{T}_p$ and $\mathcal{T}_a$ be the sets of tracklets in the primary and auxiliary cameras, respectively. As Alg. \ref{alg:tracklet} shows, we use the homography \nomenclature{$H_{p,a}$}{Homography to map the auxiliary camera image plane to that of the primary camera} $H_{p,a}$ to project detections from the auxiliary camera onto the primary camera. However, due to projective distortions, the corresponding bounding boxes in the two cameras may not necessarily overlap. Hence, we compute the optimal association cost using the Fr\'{e}chet distance \cite{frechet_two_curves} between the centroids of the detections in each tracklet as follows
%frechet distance
%\mathtt{d_f}(\tilde{\tau}_p,\tilde{\tau}_a)=\inf_{\alpha,\beta} \max_{t \in [0,1]}\left \{ d(\tilde{\tau}_p(\alpha_t), \tilde{\tau}_a(\beta_t)) \right \}
\begin{equation}
\mathcal{C}_{mc}(\tau_a,\tau_p)=\begin{cases}
\mathtt{f}(\tilde{\tau}_{p},\tilde{\tau}_{a}) & \text{if } \tilde{\tau}_a\neq \emptyset \text{, } \tilde{\tau}_p\neq \emptyset \text{, } \mathtt{f} < f_{max} \\ 
 \infty& \text{otherwise},
\end{cases}
\label{eq:cmc}
\end{equation}
where \nomenclature{$f_{max}$}{Maximum Fr\'{e}chet distance threshold that allow tracklet pairs to associate} \nomenclature{$\mathcal{C}_{mc}(\tau_a,\tau_p)$}{Optimal association cost between cross camera tracklets $\tau_p$ and $\tau_a$} $\tilde{\tau}_p$ and $\tilde{\tau}_a$ are the temporally overlapping segments of tracklets $\tau_p \in \mathcal{T}_p$ and $\tau_a \in \mathcal{T}_a$, \nomenclature{$\mathtt{f}(\tilde{\tau}_{p},\tilde{\tau}_{a})$}{Fr\'{e}chet distance between the temporally overlapping tracklet segments $\tilde{\tau}_p$ and $\tilde{\tau}_a$} $\mathtt{f}(\tilde{\tau}_{p},\tilde{\tau}_{a})$ is the Fr\'{e}chet distance of the centroids of the corresponding detections, and $f_{max}$ is the maximum distance threshold that allows tracklet pairs to be considered for association. 

We use the Hungarian algorithm again to determine optimal tracklet associations according to the costs $\mathcal{C}_{mc}(\tau_a,\tau_p)$. However, since the trajectory of a passenger that re-enters the field of view of a camera multiple times consists of a sequence of tracklets, we iteratively update the association costs until no further associations are possible. We keep track of indirectly associated tracklets by constructing the reachability graph  \nomenclature{$\mathcal{G}_{mc}$}{Tracklet reachability graph} $\mathcal{G}_{mc}=(V, E)$, which contains one edge for each pair of associated tracklets. We then set the temporal identifiers of all the tracklets in $\mathcal{T}_p$ associated with a common tracklet $\tau_a$ to the first identifier among them. That is, the temporal label of a tracklet $\tau$ is given by
\begin{equation}
l_{\tau} = \min_{(\tau_i, \tau_j) \in \mathcal{N}_{p}}{(l_{\tau_i})},
\label{eq:l_tau}
\end{equation}
where \nomenclature{$l_{\tau_i}$}{Temporal label of tracklet $\tau_i$} $l_{\tau_i}$ is the temporal label of tracklet $\tau_i$, and \nomenclature{$\mathcal{N}_p$}{Set of  tracklets connected to tracklet $\tau_p$ in the reachability graph} $\mathcal{N}_{p}$ is the set of tracklets that can be reached from tracklet $\tau_p$ on $\mathcal{G}_{mc}$, which we obtain through Depth-First Search (DFS).
%document checker view
\begin{figure}[t]
\centering
\includegraphics[width=0.95\linewidth]{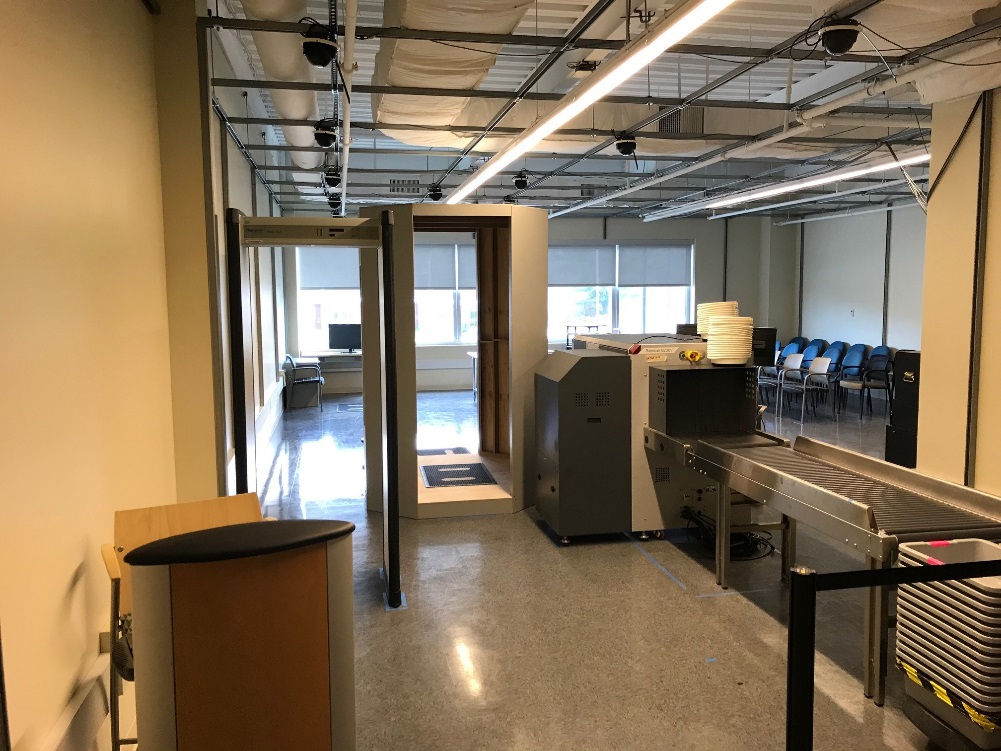}
\caption {Document checking station and divestiture area at the Kostas Research Institute simulated airport checkpoint.}
\label{fig:KRI1}
\end{figure}

\begin{figure*}[t]
\centering
\includegraphics[width=0.75\textwidth]{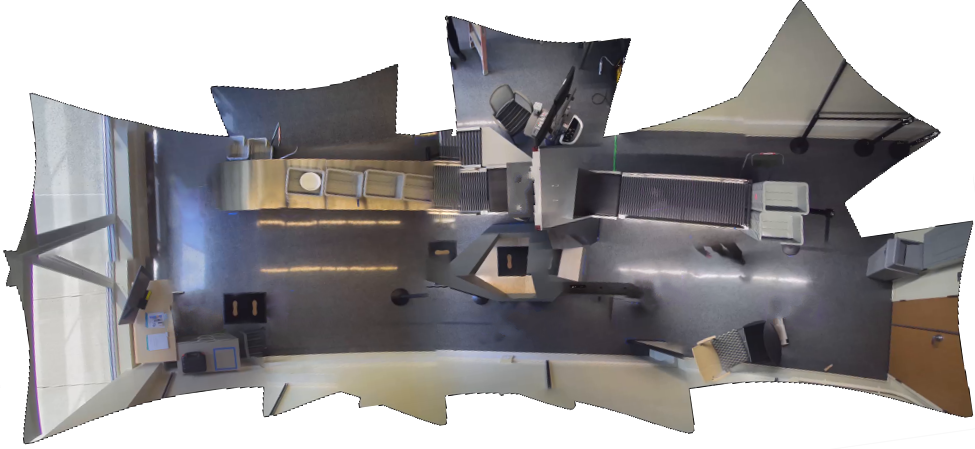} 
\caption{Panoramic overview of the camera views at the Kostas Research Institute simulated airport checkpoint.}
\label{fig:KRI_overview}
%scale=0.2
\end{figure*} 

%% results are updated
%%%%%%%%%%%%%%%%%%%%%%%%%%%%%%%%%%%%%%%%%%%%%%%%%%%%%%%%%
%%%%%%%%%%%%%%%%%%%%%%%%%%%%% RESULTS and DISCUSSIONS ####################################################
\section{Results and Discussion}
\label{sec:results}
In this section, we first discuss the datasets that we used to evaluate our algorithms. We then present an assessment of the proposed SSL approach in terms of passenger and baggage detection, followed by an evaluation of the single-camera tracking and multi-view tracklet association algorithms. Our evaluation is based on the Multi-Object Detection (MOD) and Tracking (MOT) metrics \cite{DBLP:journals/corr/Leal-TaixeMRRS15,Stiefelhagen:CLEAR_MOT}. Additional results are presented in the Supplementary Materials.

%%%%%%%%%%%%%%%% Datasets Information %%%%%%%%%%%%%%%%%%%%%%%%%%%%%%%%%%%%%%%%%%%%%%%%%%%%%%%%%%%
\subsection{Datasets}
The video datasets used in this work were recorded at the Kostas Research Institute (KRI) video analytics laboratory at Northeastern University. As shown in Fig. \ref{fig:KRI1}, the laboratory is configured to emulate a realistic airport checkpoint. It is equipped with 14 standard IP surveillance cameras (Bosch NDN-832-V03P) with $1920 \times 1080$ resolution and focal lengths between 3 mm and 9 mm. The cameras are installed approximately three meters from the floor with partially overlapping fields of view. Fig. \ref{fig:KRI_overview} shows a panoramic perspective of the fields of view of the cameras.

\begin{figure}[t]
\centering
%file size 623 KB
\includegraphics[width=\linewidth]{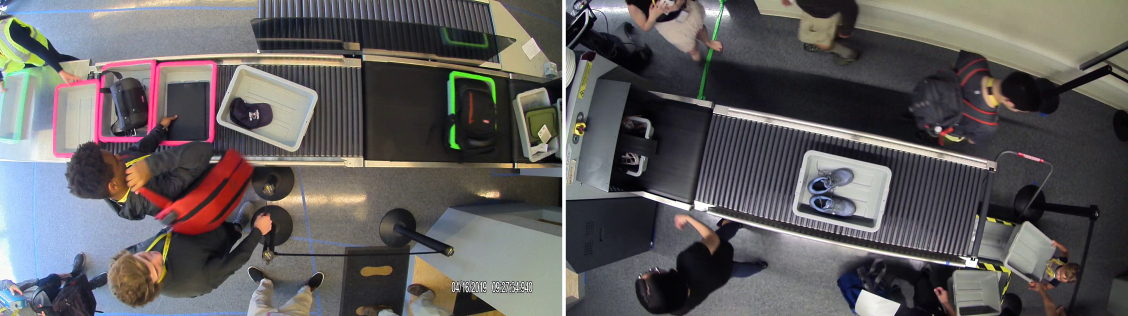} 
\caption{Sample images from the datasets collected at the  simulated airport checkpoint (left: CLASP2 and right: CLASP1 in Table \ref{tab:specifications}). The images show the divestiture area (right: camera 9) and item retrieval area (left: camera 11).}
\label{fig:KRI2}
%scale=0.2
\end{figure}

Several actors traverse the checkpoint with baggage items while performing a variety of activities commonly observed in real airports.\footnote{The datasets are available upon request at \url{alert-coe@northeastern.edu}.  Northeastern University's Institutional Review Board (IRB) and the Compliance Assurance Program Office (CAPO) within the DHS Science and Technology Directorate have reviewed the referenced human subjects research protocol and related research documentation. No compliance issues or concerns related to the use of human subjects in this protocol have been identified through the review, and the DHS policy requirements for human subjects research protocol review has been met.} These activities range from simple scenarios in which just a few passengers pass through the checkpoint in sequential order to crowded scenes in which multiple passengers divest and retrieve their items in a more erratic manner.  We collected two separate video datasets: CLASP1, which includes relatively simple scenarios, and CLASP2, which is more complex.  Fig. \ref{fig:KRI2} shows sample frames of videos from the two datasets. %In this work, we focus on videos with the highest passenger density. 
Of the 14 cameras in the laboratory, most passenger interactions take place on cameras 9 and 11. Camera 9 monitors the divestiture area and camera 11 observes the baggage retrieval area. Passengers place their belongings into bins or directly on the conveyor belt in the divestiture area. Then, after passing through the metal detector, they collect their belongings in the baggage retrieval area.

%%%%%%%%%%%% Insert Dataset Sample Image %%%%%%%%%%%%%%%%%%%%%%

\begin{table}[t]
\caption{Datasets used to evaluate our algorithms. For each video sequence, the table shows the number of passengers, baggage items, video frames, annotated frames, and the total number of annotated bounding boxes.}
%Labels / Frames
\label{tab:specifications}
\centering
\setlength\tabcolsep{4pt}
\begin{tabular}{lcccccc}
\toprule
\multirow{2}{*}{Dataset} &Video  &Pass- & Bag-     &Video & Annotated &Bounding  \\ 
& seq.   & engers & gage     & frames & frames/rate [fps] &  boxes\\
\midrule
\multirow{5}{*}{CLASP1} &A       & 12       & 10        & 6,030      &288 (1)     &~~995 \\
&B       &  12     & 10        & 6,180      & 564 (2)       & 1,720 \\
&C       & 8     & 9         & 6,030      & 491 (2)       & ~~853 \\ 
&D       & 12     & 8         & 6,030      & 523 (2)       & 1,197  \\
&E       & 9     & 9         & 4,719      & 1,648 (10)       & 4,254 \\ 
\hline

\multirow{3}{*}{CLASP2}
&F       &20     &20        &12,910      &179 (0.01)       & 737 \\
&G       & 38     &31       &10,390      & 1,346 (3)        & ~4,826 \\ 
&H       & 35     & 29      &11,200       & 198 (0.01)      & 900  \\
\hline
Total & -- & 146    & 126     &63,489     & 5,237              & 15,482\\
\bottomrule
\end{tabular}
\end{table}

As Table \ref{tab:specifications} shows, a total of 146 passengers carrying 126 baggage items leave and re-enter the fields of view of the cameras several times.   We manually annotate the videos with uniquely identified axis-aligned bounding boxes. Given the large number of video frames available in the datasets, the annotation rate for the video sequences varies between $0.01$ and $10$ frames per second (fps). We randomly partition each dataset into a training set containing $80\%$ of the video frames and a test set with the remaining $20\%$. For a fair comparison, the Supervised Learning (SL) and SSL models are trained using only the frames from the training set, but the SSL models are fully self-supervised and do not use any manual annotations.  However, disregarding every video sequence that includes annotated frames would substantially limit the amount of data available for the computation of tracking performance measures. Hence, to assess tracking performance, we consider all the annotations listed in Table \ref{tab:specifications}. The only method that uses the training set annotations is the SL approach. Although this evaluation strategy favors that method, it also more accurately reflects the generalization performance of the SSL approaches to unseen data.  Due to space limitations, the results presented in this section were obtained using the aggregated CLASP1 and CLASP2 test sets. Dataset-specific results are given in the Supplementary Materials.

%training progression curve: SSL Models
\begin{figure}[t]
\centering
\includegraphics[trim=0.3 0 0 0, clip, height=1.48in]{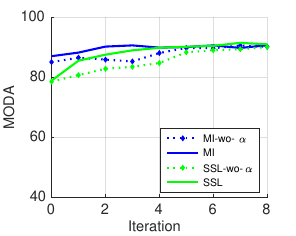}
\includegraphics[trim=23 0 0 0, clip, height=1.48in]{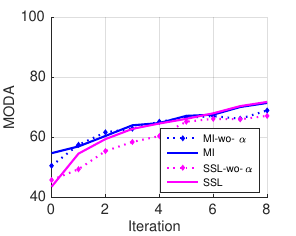}
\caption {MODA measures for person (left) and baggage (right) classes during SSL training.}
\label{fig:SSL_progress}
\end{figure} 

%%%%%%%%%%% PR curve  %%%%%%%%%%%%%%%%%%%%%%%%%%%
\subsection{Self-Supervised Learning Detection Performance}
\label{sec:ssl-detection-performance}

During training, we freeze the network weights up to the region proposal network layer so that the pre-trained backbone features are effectively used in the downstream task. We use an initial learning rate of $5\mathrm{e}{-3}$, mini-batch size per image $N=256$,  $r=20$ different orientations, and a cluster confidence threshold $\lambda=0.1$. Similar to the baseline model, we use stochastic gradient descent  with a momentum of $0.9$, weight decay of $1\mathrm{e}{-4}$. At each SSL iteration, we fine-tune the model for $20$k iterations,  reducing the learning rate by a factor of $10$ at every $5$k iterations. In our evaluation, we use an IoU threshold of $0.5$, and a non-maximum suppression threshold $\eta_{\text{nms}}=0.3$ for all the models. The detection threshold for region proposal generation is $\eta_{\text{det}}=0.5$.

\begin{figure}[t]
\centering
\includegraphics[trim=0.5 0 0 0, clip, height=1.47in]{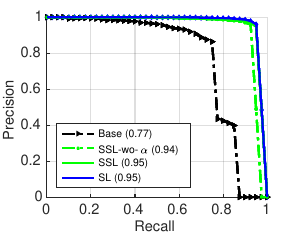}
\includegraphics[trim=20 0 0 0, clip, height=1.47in]{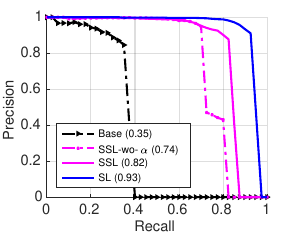}
\caption {Precision-recall curves for person (left) and baggage (right) detection.
%using the SL and SSL models. 
The legend shows the average precision of the models.}
\label{fig:PR}
\end{figure} 

Fig. \ref{fig:SSL_progress} shows the Multi-Object Detection Accuracy (MODA) of our model as a function of the number of SSL iterations. To illustrate the impact of the cluster confidence score, we also evaluate a model in which the samples are not weighed by their scores (SSL-wo-$\alpha$). Instead, this model uses a hard threshold $\lambda \leq 0.4$ to discard noisy detections during training. The figure also shows the performance of the Multiple-Inference (MI) strategy used to generate the pseudo-labels, which reflects the quality of the pseudo-labels before SSL training. That is, in the MI model, the pseudo-labels themselves are used as model predictions. As the figure indicates, the SSL models gradually approach the performance of the MI strategy.  The incorporation of cluster confidences not only increases the speed of convergence of the models but also leads to noticeable performance gains, particularly for baggage items. 

Fig. \ref{fig:PR} shows the precision-recall curves for passenger and baggage detection using four detector models: pre-trained PANet (baseline), PANet trained using SL, SSL-wo-$\alpha$, and SSL. 
%All the results are based on an IoU threshold of $0.5$, which allows for the correct detection of passengers despite the substantial variability in bounding box size as passengers change their orientations or as they move their arms to interact with baggage items. 
Even though the SSL models are trained without manual annotations, 
%Fig. \ref{fig:PR} shows that it increases the area under the curve for both datasets by a large margin in comparison with the baseline model.
%, leading to a maximum of 94\%, 77\% for person and bag class in CLASP1 and 92\%, 80\% in CLASP2. %, and the number of rotation angles is $r=20$ in pseudo-label generation step. 
%The SSL models 
they perform on par with the SL model for passengers. 
For baggage items, the maximum average precision for the baseline model is less than half of the performance of the SSL models. 
As illustrated in Fig. \ref{fig:bag_failure_cases}, the performance difference between the SL and SSL models is due to two main issues: i) appearance similarities among bags and certain garments/items  placed inside security bins, and ii) baggage items that can only be partially observed before being placed on the conveyor belt. 

%% Bag detection failure cases
\begin{figure*}[h]
\centering
%file size 1.8 MB
\includegraphics[width=0.99\textwidth]{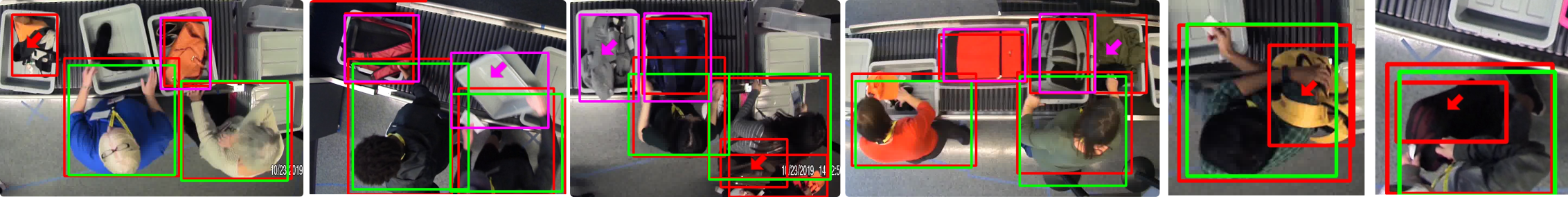}
\centering
\caption {Sample results showing failure cases for baggage detection using the SSL model in the CLASP2 dataset. The magenta arrows indicate bag-like object detections that are not annotated (false positives), the red arrows indicate annotated baggage items the model fails to detect (false negatives), the green bounding boxes show passenger detections, and the red bounding boxes represent the manual annotations for both classes.}
\label{fig:bag_failure_cases}
\end{figure*}
%%%%%%%%%%%%%%%%%%%%%%%%%%%%%%%%%%%%%%%%%%%%%%%%%%%%%%%%%%%%%%%%

%MOD evaluation measure for person and bag classes
\begin{table*}[t]
\caption{Passenger and baggage detection evaluation.}
\label{tab:detection_person_bag}
%\begin{figure}[t]
\centering
\setlength{\tabcolsep}{5pt} % Default value: 6pt
\footnotesize
\begin{tabular}{lcccccccccccccc}
\hline        
\multicolumn{1}{l}{Model} &
\multicolumn{2}{c}{Method} &
\multicolumn{2}{c}{$\uparrow$Rcll} & \multicolumn{2}{c}{$\uparrow$Prcn} & \multicolumn{2}{c}{$\uparrow$TP} &\multicolumn{2}{c}{$\downarrow$FP} &\multicolumn{2}{c}{$\downarrow$FN} &\multicolumn{2}{c}{$\uparrow$MODA}\\
         &$\alpha$ &reg.              & person          & bag         & person         & bag         & person         & bag    & person         & bag     & person         & bag  & person         & bag  \\ 
\hline
Baseline &\xmark &\xmark &73.8 &37.1    &87.0 &82.9    &1560 &426    &228 &85   &552 &724   &62.8 &29.3 \\ 
$\text{SSL}$ &\xmark &\xmark &93.8 &73.8    &92.3 &82.5    &1989 &858  &155 &194   &123 &291   &86.0 &57.6 \\
$\text{SSL}$ &\checkmark &\xmark &93.5 &75.9   &93.5 &86.1   &1985 &863   &134 &144   &127 &286  &87.1 &62.9 \\
$\text{SSL}$ &\xmark &\checkmark &93.6 &73.1  &\underline{96.2} &\underline{92.5}  &1985 &844  &\underline{73} &\bf{71}   &127 &305  &90.1 &67.1 \\
$\text{SSL}$ &\checkmark &\checkmark &\bf{95.7} &\underline{78.6}   &96.0 &91.8   &\bf{2025} &\underline{903}   &79 &83   &\bf{87} &\underline{246}   &\underline{91.8} &\underline{71.5} \\
$\text{SL}$ &\xmark &\xmark &\underline{95.6} &\bf{91.4}   &\bf{96.4} &\bf{92.8}  &\underline{2022} &\bf{1048}   &\bf{70} &\underline{83}   &\underline{90} &\bf{101}  &\bf{92.1} &\bf{84.4} \\
\hline
\end{tabular}
\end{table*}

%det models test performance comparison
\begin{figure*}[t]
\centering
%file size 478 KB
\includegraphics[width=0.99\textwidth]{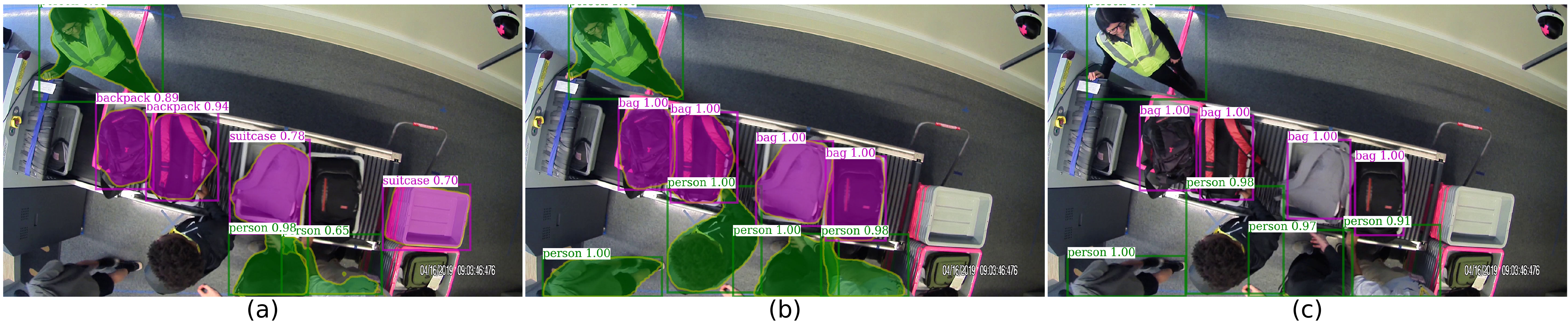}
\centering
\caption {Qualitative detection results  on the CLASP2 dataset using (a) Baseline, (b) SSL, and (c) SL models (the SL model only predicts bounding boxes).}
\label{fig:det_models_test_compare}
\end{figure*}

Table \ref{tab:detection_person_bag} demonstrates the benefits of incorporating cluster uncertainties in the SSL loss function (column $\alpha$) and of the proposed cluster regression technique (column \emph{reg.}). The method that incorporates both cluster uncertainty and regression is equivalent to the approach identified as SSL in Figs. \ref{fig:SSL_progress} and \ref{fig:PR} whereas the method that does not include cluster confidences corresponds to SSL-wo-$\alpha$. The results in the table correspond to the point that maximizes the $F_1$ score of the curves in Fig. \ref{fig:PR} at the best performing SSL iteration. The top-performing method in Table \ref{tab:detection_person_bag} and in the remainder of this section is highlighted in boldface, the second-best is underlined, and ties are broken according to the MODA/MOTA results.

In comparison with the baseline model, our SSL algorithm substantially increases the recall (Rcll) and precision (Prcn) for passenger detection, which is a result of  improvements in true positive (TP), false positive (FP), and false negative (FN) detections. The cluster confidence scores substantially reduce the contribution of low-confidence pseudo-labels, especially for baggage items, leading to a noticeable increase in the number of true positives. Cluster regression corrects pseudo-label errors caused by inaccurate bounding boxes generated from poor segmentation results. As a result, the reduction in false positives for both classes is even more pronounced when cluster regression is incorporated.
Overall, our SSL framework shows a relative MODA score improvement of 46\% for passengers and 144\% for baggage items with respect to the baseline model. 

Fig. \ref{fig:det_models_test_compare} shows qualitative results for the models under consideration. In comparison with the SL model, the SSL models not only improve the accuracy of the predicted bounding boxes but also generate improved segmentation masks since they are trained using instance segmentation pseudo-labels.
%%%%%%%%%%%%%% MOT Quantitative Tracking Results
%Note: Since the difference between the detections and tracking results in CLASP1 and CLASP2 are significant, we will demonstrate the camera-wise tracking results for CL1 and Cl2 in separate bar graph.
\begin{table*}[t]
\begin{minipage}{1\textwidth}
\centering
\caption{Single-camera tracking evaluation for person and baggage classes.}
\label{tab:track_evaluation}
\setlength\tabcolsep{5pt}
\begin{tabular}{l|lccccccccccccccccccc}
\toprule
                 Class     &Model &$\alpha$
                 &SCA &GT
                 &$\uparrow$IDF1 &$\uparrow$IDR &$\uparrow$IDP & $\uparrow$Rcll & $\uparrow$Prcn  & $\downarrow$FP & $\downarrow$FN & $\uparrow$MT & $\downarrow$ML &  $\downarrow$IDs & $\downarrow$FM  & $\uparrow$MOTA & $\uparrow$MOTP \\ \midrule

\multirow{5}{*}{Person}  &Baseline &\xmark &\xmark &391 &84.5 &83.3 &86.1  &91.2  &94.6 &750 &753 &283 &42 &93 &152 &84.0 &85.5\\
                         &$\text{SSL}$ &\xmark &\xmark  &391 &87.8 &87.2 &88.3  &95.1  &96.4 &\bf{350} &554 &319 &31 &80 &123 &90.1 &85.2 \\
                         &SSL &\checkmark &\xmark &391 &\underline{88.4}  &\underline{87.9} &\underline{88.9} &\underline{95.6}  &\underline{96.6} &\underline{354} &\underline{438} &326 &27 &86 &\underline{122} &\underline{90.7} &\underline{85.2} \\
                         &SSL &\checkmark &\checkmark &391 &\bf{88.5}  &\bf{88.1} &\bf{88.9} &\bf{95.6}  &96.5 &358 &\bf{435} &\underline{326} &\underline{26} &\bf{76} &123 &\bf{90.8} &\bf{85.2} \\
                           &SL &\xmark &\xmark &391 &87.0  &86.4 &87.7 &95.2  &\bf{96.9} &357 &457 &\bf{332} &\bf{24} &\underline{86} &\bf{121} &90.5 &85.2\\
                        \cline{1-18}
                       
%%% Bag A-exp9A                        
\multirow{5}{*}{Bag}   &Baseline &\xmark &\xmark &255 &67.5  &57.0 &\underline{86.4} &61.1  &92.3  &431 &1800 &108 &73 &\bf{31} &\bf{89} &54.3 &\bf{81.0} \\
                     &SSL &\xmark &\xmark &255 &78.9  &74.2 &85.4 &81.0  &\bf{93.7}  &\bf{308} &1014 &159 &38  &71 &105 &72.9 &80.4\\
                     &SSL &\checkmark &\xmark &255 &81.3  &78.2 &85.5 &84.4  &92.3  &401 &822 &169 &28  &68 &97 &75.1 &80.4\\
                     &SSL &\checkmark &\checkmark &255 &\underline{84.3}  &\underline{81.1} &\bf{88.5} &\underline{84.4}  &\underline{92.3}  &401 &\underline{822} &\underline{169} &\underline{28} &\underline{48} &97 &\underline{75.6} &\underline{80.4} \\
                     &SL &\xmark &\xmark &255 &\bf{85.3}  &\bf{86.6} &84.1 &\bf{94.7}  &91.7  &\underline{387} &\bf{339} &\bf{226} &\bf{10}  &103 &\underline{69} &\bf{83.2} &80.0\\
\bottomrule
\end{tabular}
\end{minipage}
\end{table*}

\subsection{Single-Camera Tracking}
\label{sub:tracking}
%To evaluate the performance of SSL detector based person and bag tracking, we provide the SSL model in the tracktor \cite{tracktor_2019_ICCV} to detect, regress, and generate the tracklets  at 30 frames per second inputs. 
We compare the performance of our single-camera tracking algorithm using the proposed SSL detectors with the pre-trained baseline detector and the SL detector. We also evaluate the impact of our SCA algorithm, described in Section \ref{sec:sca}, where we use $t_{th}=3$ seconds and $\delta_{max}=200$ in Eq. \ref{eq:single_tracklet}. To preserve the entirely self-supervised nature of our pipeline, we refrain from fine-tuning the re-identification module of the baseline tracker, which is pre-trained on the MOT17 \cite{DBLP:journals/corr/Leal-TaixeMRRS15} dataset. 
To dissociate the evaluation of the tracking method from our MCTA approach, we use a modified version of the annotations in Table \ref{tab:specifications} where a passenger that re-enters the field of view of a camera receives a new identifier. Thus, the number of unique ground truth passenger identifiers (column GT in Table \ref{tab:track_evaluation}) is much higher than those listed in Table \ref{tab:specifications}. We evaluate our system's ability to maintain consistent passenger identifiers across multiple perspectives in Section \ref{sub:mcta_resuls_discussion}.

%%%%%%%%%%% Insert Sample PxBy tracking results: cross camera and PxBx %%%%%%%%%%%%%%%%%%%%%
\begin{figure*}[h]
\centering
%file size 972 KB
\includegraphics[width=\textwidth]{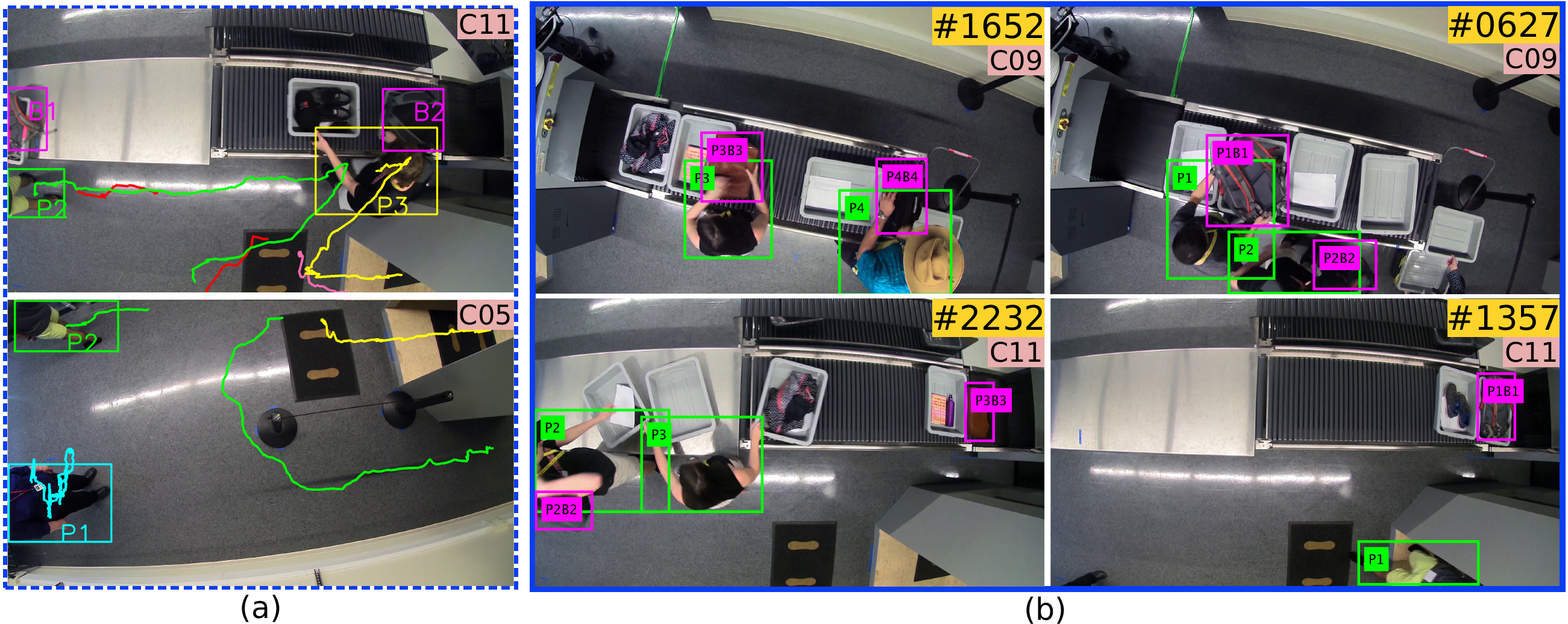}
\centering
\caption {Sample results showing (a) cross-camera passenger association between cameras 5 and 11 using MCTA, and (b) tracking and association between passengers and baggage items  where the top and bottom rows show image sequences from cameras 9 and 11 respectively. We associate passenger tracklets in cameras 9 and 11 by leveraging the associations between cameras 2 and 5 (passengers flow in Fig. \ref{fig:KRI_overview}: C9$\rightarrow$C2$\rightarrow$C5$\rightarrow$C11). Baggage items are associated using temporally constrained distance-based matching when each item receives a unique identifier $PiBj$, representing the $j$-th item from the $i$-th passenger.}
\label{fig:track}
\end{figure*} 

As Table \ref{tab:track_evaluation} shows, the SSL-wo-$\alpha$ and SSL approaches outperform the tracker using the baseline detector by a large margin.  The notable improvements in identity-based $F_1$ (IDF1), recall (IDR), and precision (IDP) \cite{IDF1_scor_eccv} as well as in standard recall and precision are primarily a result of the reduction in false positives and false negatives generated by the SSL model. Self-supervision also improves the tracking-specific metrics of mostly tracked (MT), mostly lost (ML), identity switches (IDs), and fragmented (FM) trajectories \cite{Stiefelhagen:CLEAR_MOT}. As a result, our method produces substantial gains in MOTA. Again, both SSL models perform on par with the SL model for person tracking. For baggage items, we see similar performance improvements, but the challenges illustrated in Fig. \ref{fig:bag_failure_cases} again preclude the SSL models from reaching the performance of the SL strategy. Finally, our SCA algorithm leads to further performance gains, particularly in terms of IDs.

\subsection{Multi-Camera Tracklet Association}
\label{sub:mcta_resuls_discussion}
%% Multi-camera Association Table:
%NOTE: Due to the mixed results using velocity, current evaluation is based on the centroids only. 

%Improved ID measurement using tracklets association: Top row: A-9A, bottom row:D-10A
\begin{table}[t]
\centering
\caption{MCTA evaluation. The column labeled Dist.  indicates whether we employ the Hausdorff ($d_h$) or Fr\'{e}chet ($d_f$) distance to evaluate tracklet similarity.}
\label{tab:ID_Measure_Tracklet_Asso}
\setlength\tabcolsep{2.2pt}
\begin{tabular}{ccccc|ccccc}
\toprule
%we found maximum number of re-entry in camera 2 from 9 (out of scope for evaluation): #re-entry: camera 2 to 9 and camera 5 to 11. Here, #re-entry from camera 9 to 2 and 11 to 5 are not counted to show in evaluation table.
Dist. &SL &SSL-wo-$\alpha$ &SSL &MCTA  & $\uparrow$IDF1  & $\uparrow$IDR  & $\uparrow$IDP  & $\downarrow$IDs  & $\uparrow$MOTA            \\
 
\hline
  \multirow{3}{*}{-}   &\xmark &\checkmark &\xmark  &\xmark &82.1  &83.2  &81.0  &157  &88.2    \\
                        &\xmark &\xmark &\checkmark  &\xmark &82.0  &83.1  &80.9  &170  &88.5   \\
                        &\checkmark &\xmark &\xmark  &\xmark &81.5  &82.8  &80.4  &170  &88.0    \\
                    \cline{1-10} 
                      \multirow{3}{*}{$d_h$}
                                         &\xmark &\checkmark &\xmark &\checkmark  &87.4  &88.9  &86.4  &\underline{115}  &88.8    \\
                                         &\xmark &\xmark &\checkmark &\checkmark  &84.8  &88.6  &86.2  &140  &89.0    \\ 
                 &\checkmark &\xmark &\xmark &\checkmark  &86.2  &87.5  &85.0  &134  &88.6    \\
 \cline{1-10}
                    \multirow{3}{*}{$d_f$}
                                         &\xmark &\checkmark &\xmark &\checkmark  &\underline{88.0}  &\underline{89.3}  &\underline{86.8}  &\bf{115}  &88.9   \\
                   
                                        &\xmark &\xmark &\checkmark &\checkmark  &\bf{88.2}  &\bf{89.5}  &\bf{87.0}  &132  &\bf{89.1}    \\
                                        &\checkmark &\xmark &\xmark &\checkmark  &86.7  &88.1  &85.5  &122  &\underline{89.0}    \\
\bottomrule
\end{tabular}
\end{table}

We evaluate the performance of our MCTA algorithm using the same experimental procedure described in the previous section, with the exception that passengers are now assigned unique identifiers as they leave and re-enter the fields of view of the cameras. 
Based on the overall flow of passengers through our simulated checkpoint, cameras 9 and 11 are the primary cameras for our tracklet association method (Alg. \ref{alg:tracklet}). Cameras 2 and 5, the cameras immediately below them in Fig. \ref{fig:KRI_overview}, are the respective auxiliary cameras. For a fair comparison among the detectors, we generate tracklets in the auxiliary cameras using the corresponding SL or SSL model used in the primary cameras (i.e., trained using only frames from the primary camera). To provide a set of reference performance measures, we first evaluate our tracking algorithms in the absence of a MCTA mechanism. We then assess the performance of our association method when tracklet similarity is computed using the Fr\'{e}chet distance and the more traditional Hausdorff distance \cite{tracklet_asso_hausdorff} with $f_{max}=0.25$ in Eq. \ref{eq:cmc}. 

Table \ref{tab:ID_Measure_Tracklet_Asso} shows that tracklet association improves the IDF1 measure by up to $6.2\%$.
%, with an average improvement over all the datasets of approximately $3.9\%$, $4.4\%$, and $3.5\%$ in IDF1, IDR, and IDP, respectively. 
This is mainly a consequence of the dramatic reduction in the number of identity switches. Using the Fr\'{e}chet distance to determine tracklet similarity provides consistent performance improvements in all the metrics under consideration, especially for the SSL strategy.
%For example, in Dataset CL2, no passenger re-enters the field of view of camera 9, and therefore no improvement in identity-based metrics can be obtained. 
The more modest gains in MOTA (up to $1.0\%$) demonstrate the need for measures that focus specifically on the impact of identity switches on tracking performance. 
%Again, the SSL-$\alpha$ detector based MCTA outperforms the other models in both datasets except for a few additional identity switches. 

Fig. \ref{fig:track}(a) illustrates the tracklet association procedure between cameras 5 and 11. As the passengers with identities P2 and P3, whose trajectories are represented in green and yellow, move from the field of view of camera 5 to camera 11, their tracklets are projected from the former camera to the latter. The projected trajectories (red for P2 and pink for P3) are successfully associated with the tracklets from camera 11 based on the Fr\'{e}chet distances among their temporally overlapping segments. In the instant shown in the figure, passenger P2 is re-entering the field of view of camera 5, and the corresponding tracklet is also correctly associated with that passenger's tracklet in camera 11. Hence, the passenger's identity is successfully handed off between the cameras. Fig. \ref{fig:track}(b) demonstrates a potential application of the proposed system. Baggage items are associated with passengers when they are divested in camera 9 and their identifiers can be verified at retrieval time, which is observed in camera 11.

\section{Conclusion}
\label{sec:conclusion}
We propose a multistage tracking-by-detection framework to overcome performance limitations of object detection and tracking algorithms in overhead camera videos for which limited training data is available. Our framework is composed of an SSL mechanism to fine-tune object detection models to specific camera views without the need for manual annotations and an MCTA method that only requires the homographies among neighboring cameras. Our experiments show that the proposed framework can accurately detect and track passengers and baggage items across camera views in airport checkpoint scenarios. Our framework is flexible and scalable. It requires no training data,  incurs no detection computational overhead at inference time, and is independent of the number of cameras in the network.

Our framework also allows seamless integration of additional data augmentation strategies and of manually annotated data when it is available. Our experiments show that these strategies further improve the selectivity of our detector, particularly for baggage items. For simplicity, our association methods are performed offline, i.e., after all the tracklets have been generated. However, it would be simple to implement online versions of the algorithms since the single-view association method can be executed whenever a new trajectory is initiated and each iteration of the MCTA algorithm can be performed once a trajectory in an auxiliary camera is terminated. The implementation of a real-time version of our tracklet association method is part of our future work.

\bibliographystyle{IEEEtran}
\bibliography{IEEEabrv,egbib}

\title{\LARGE Supplementary Materials: Tracking Passengers and Baggage Items using Multiple Overhead Cameras  at Security Checkpoints}%
% Authors at the same institution
\author{Abubakar Siddique,~\IEEEmembership{Student Member,~IEEE,}
          Henry Medeiros,~\IEEEmembership{Senior Member, IEEE}% <-this % stops a space
%\thanks{Department of Electrical and Computer Engineering, Marquette University, Milwaukee, WI, 53233, e-mail: abubakar.siddique@marquette.edu \hspace{2cm} henry.medeiros@marquette.edu.}% <-this % stops a space
% \thanks{This material is based upon work supported by the U.S. Department of Homeland Security, Science and Technology Directorate, Office of University Programs, under Award Number 2013-ST-061-E0001-04. The views and conclusions contained in this document are those of the authors and should not be interpreted as necessarily representing the official policies, either expressed or implied, of the U.S. Department of Homeland Security.}
% \thanks{Manuscript received January DD, 2022; revised Month DD, 2022.}
}

% The paper headers
\markboth{IEEE Transactions on Systems, Man, and Cybernetics: Systems,~Vol.~X, No.~Y, December 2022}{Siddique and Medeiros: Tracking Passengers and Baggage Items using Multiple Overhead Cameras at Security Checkpoints}

\maketitle
%\ifwacvfinal\thispagestyle{empty}\fi
%To develop an automated system, ALERT (Awareness and Localization of Explosives-Related Threats) starts CLASP (Correlating Luggage and Specific Passengers) project supported by DHS Science and Technology Directorate through the DHS Office of University Programs. A video analytics lab at Northeastern University collect video data of passenger moving through a mock airport security checkpoint to simulates the real-world conditions. The goal of CLASP project is to assist Transportation Security Administration (TSA) for detecting security incidents such as theft of items, bags left behind the scene, or suspicious behavior of the passengers. 

%%%%%%%%% ABSTRACT
\begin{abstract}
This document supplements our main paper with additional experimental results on the CLASP1 and CLASP2 datasets. We extend our Self-Supervised Learning (SSL) approach into a Semi-Supervised Learning (Semi-SL) mechanism to further improve target detection performance, especially for baggage items. We also investigate the impact of additional data augmentation strategies, rotation resolution, and the computational requirements of our proposed technique. These additional evaluation results show that our algorithm outperforms the baseline as well as state-of-the-art supervised and semi-supervised approaches. 
\end{abstract}
\begin{IEEEkeywords}
Detection, Tracking, Association, Homography, Tracklet, Multi-camera, Surveillance.
\end{IEEEkeywords}
%single view detection performance

%Single camera SSL detections comparisons
\begin{figure*}[h]
\centering
\begin{minipage}{0.9\linewidth}
\centering
\includegraphics[width=0.32\textwidth,height=1.85in]{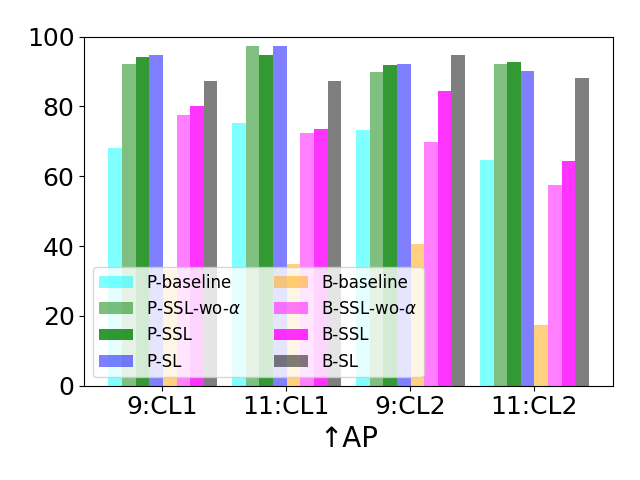}
\includegraphics[width=0.32\textwidth,trim=60 0 0 0, clip,height=1.85in]{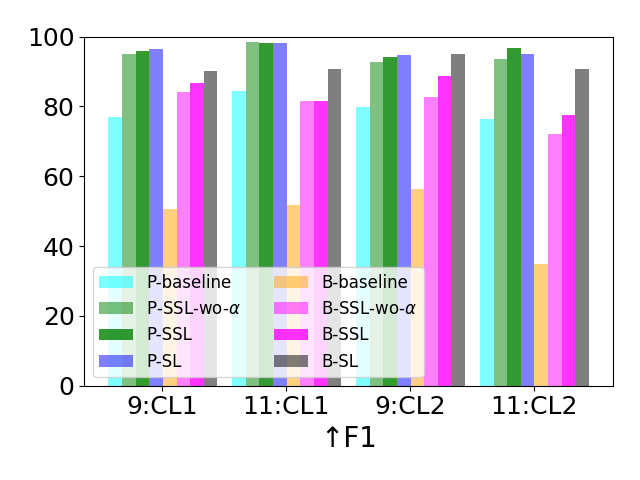} 
\includegraphics[width=0.32\textwidth,trim=60 0 0 0, clip,height=1.85in]{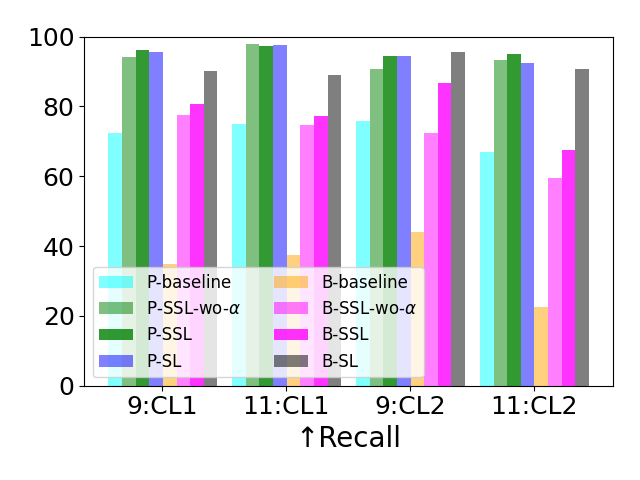} \\
\includegraphics[width=0.32\textwidth,height=1.85in]{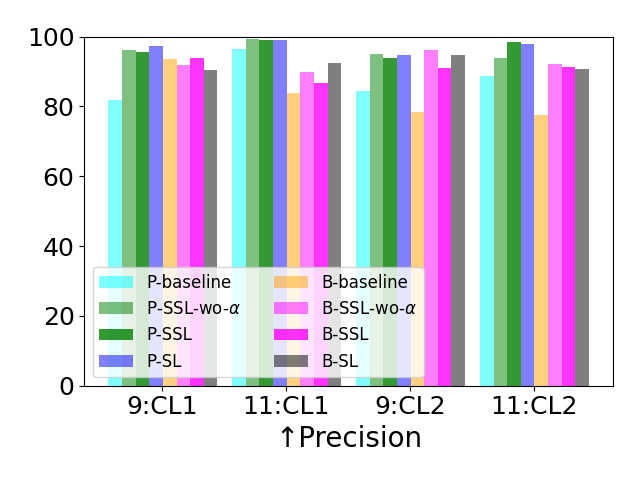}
\includegraphics[width=0.32\textwidth,trim=60 0 0 0, clip,height=1.85in]{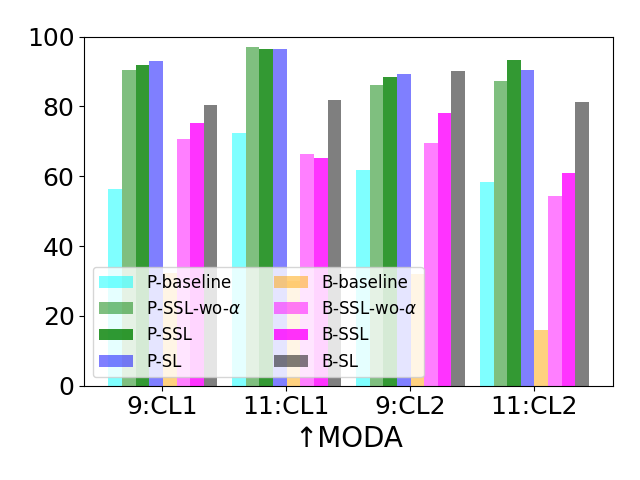}
\includegraphics[width=0.32\textwidth,trim=60 0 0 0, clip,height=1.85in]{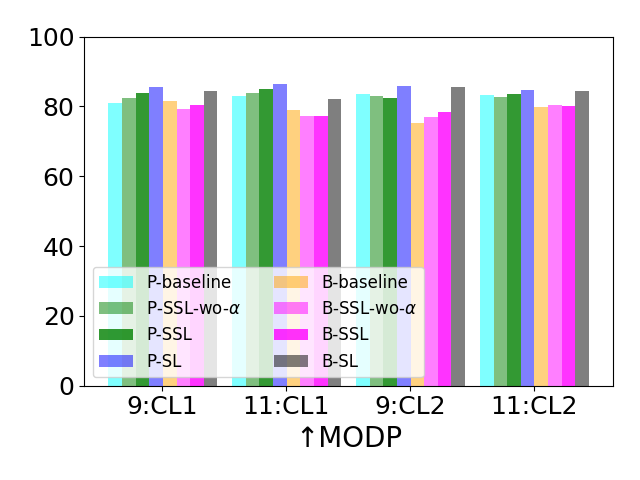}
\caption {Passenger and baggage detection performance in cameras 9 and 11 for the CLASP1 (CL1) and CLASP2 (CL2) datasets.  Here, P stands for passenger and B for baggage. A description of the methods under consideration is given in Section IV.B of the main paper.}
\label{fig:det_bar}
\end{minipage}
\end{figure*}

\section{Single Camera Detections}
This section presents a breakdown on the performance of our SSL detector for individual cameras in the CLASP1 and CLASP2 datasets. It also evaluates the performance impact of additional data augmentation strategies, number of rotation angles used for data augmentation, and incorporation of labeled data in a semi-supervised approach.
\label{sec:sc-detection}

\subsection{Self-Supervised Learning} 
Fig. \ref{fig:det_bar} shows a detailed breakdown of the performance of our SSL detection model for individual camera views in the CLASP1 and CLASP2 datasets. The high recall, precision, and MODA values indicate that our SSL approach detects most passengers correctly in these video sequences. The average precision (AP) for passenger detection is slightly higher for camera 11 in both datasets. The main factor contributing to this performance difference is that in camera 9, passengers are only partially visible most of the time, whereas camera 11 has a better view of the region where the passengers stand next to the conveyor belt. On the other hand, this also contributes to the lower baggage detection performance in camera 11. That is, in camera 11, partially observed baggage items being carried by passengers (see Fig. \ref{fig:bag_failure_cases}) are much more common than in camera 9. As with passenger detection, we observed similar baggage detection improvements in the camera-specific performance comparisons. This performance could be further improved by using additional unlabelled video frames available in the CLASP1 and CLASP2 datasets to train the SSL models.

%% Bag detection failure cases
\begin{figure*}[h]
\centering
%file size 1.8 MB
\includegraphics[width=0.98\textwidth]{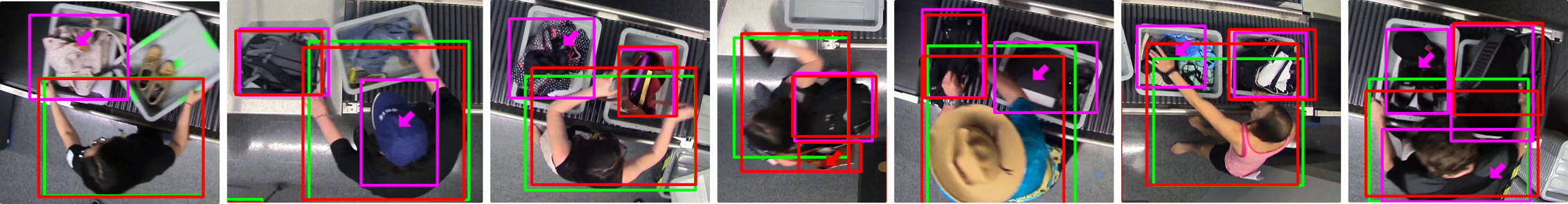}
\centering
\caption {Additional illustrative failure cases for baggage detection using the SSL model in the CLASP1 dataset (see Fig. 10 in the main paper for failures in the more challenging CLASP2 dataset). The magenta arrows indicate bag-like object detections that are not annotated (false positives), the red arrows indicate annotated baggage items the model fails to detect (false negatives), the green bounding boxes show passenger detections, and the red bounding boxes represent manual annotations for both classes.}
\label{fig:bag_failure_cases}
\end{figure*}
%We found similar performance improvements in the CLASP1 dataset. 
%In this experiment, the pre-trained models of \cite{panet_liu2018} and \cite{SoftTeacher_2021_iccv} trained on COCO and we found similar quantitative evaluation when they use full set of manual annotations. 

% brightness factor is chosen uniformly from [max(0, 1-brightness, 1+brightness] and hue is chosen uniformly from [-hue, hue]
\subsection{Additional Data Augmentation Strategies}
We investigate the impact of other data augmentation strategies  during SSL training, including color jittering and motion blur along with multiple rotations. For color jittering, we increase/decrease image brightness, contrast, saturation, and hue by a factor sampled uniformly from the range $\left[0, \text{max}_\text{jit}\right]$, where $\text{max}_\text{jit}$ is $0.4$ for brightness, $0.5$ for contrast, $0.2$ for saturation, and  $0.05$ for hue. To emulate motion blur, we use Gaussian blur with kernel size uniformly sampled from the set $\left\{5,\ldots,9\right\}$ and standard deviation sampled from the interval $\left[0.1, 5\right]$.   We observe that applying color jittering and motion blur on the pseudo-label augmentation further improves MODA scores by up to  $2.9\%$ and $4.8\%$ for passengers and baggage items, respectively. For a fair comparison, we reduced the number of rotation angles used for augmentation such that the total number of augmented images remains the same in both scenarios. Maintaining the original number of rotations would further increase performance gains.

%%%% person
\begin{table}[h]
%with rotation+mixed aug: SSL base to iter0
%%%%%CLASP1
% Average Precision: 0.8916
%  F1  Rcll  Prcn|  FAR     GT     TP     FP     FN| MODA  MODP 
% 91.5  92.4  90.5| 0.02   1200   1109    116     91| 82.8  83.2
%bag
% Average Precision: 0.4836
%  F1  Rcll  Prcn|  FAR     GT     TP     FP     FN| MODA  MODP 
% 64.6  48.9  95.2| 0.00    644    315     16    329| 46.4  80.5 

%%%%%CLASP2
%bag
% Average Precision: 0.4554
%  F1  Rcll  Prcn|  FAR     GT     TP     FP     FN| MODA  MODP 
% 61.9  46.1  94.0| 0.00    505    233     15    272| 43.2  77.1 
%person
% Average Precision: 0.8421
%  F1  Rcll  Prcn|  FAR     GT     TP     FP     FN| MODA  MODP 
% 88.0  86.3  89.8| 0.00    912    787     89    125| 76.5  84.0 

%%%with rotation only
%%%%%CLASP1
%bag
% Average Precision: 0.4344
%  F1  Rcll  Prcn|  FAR     GT     TP     FP     FN| MODA  MODP 
% 59.7  43.2  96.5| 0.00    644    278     10    366| 41.6  80.8
%person
% Average Precision: 0.8927
%  F1  Rcll  Prcn|  FAR     GT     TP     FP     FN| MODA  MODP 
% 92.0  92.2  91.8| 0.02   1200   1106     99     94| 83.9  83.2

%%%%%CLASP2
%bag
% Average Precision: 0.4742
%  F1  Rcll  Prcn|  FAR     GT     TP     FP     FN| MODA  MODP 
% 62.5  48.1  89.0| 0.00    505    243     30    262| 42.2  77.0 
%person
% Average Precision: 0.7942
%  F1  Rcll  Prcn|  FAR     GT     TP     FP     FN| MODA  MODP 
% 86.2  82.3  90.4| 0.00    912    751     80    161| 73.6  84.1

\centering
\setlength\tabcolsep{2pt} % default value: 6pt
%\begin{minipage}{1\textwidth}
\caption{Performance impact of additional data augmentation strategies in the SSL iterations.}
\label{tab:aug_ablation} 
\begin{tabular}{c|ccccccccc}
\toprule
\multicolumn{1}{c}{Dataset} &
\multicolumn{3}{c}{Method} &
\multicolumn{2}{c}{$\uparrow$AP} & \multicolumn{2}{c}{$\uparrow$ $F_1$} & \multicolumn{2}{c}{$\uparrow$MODA}\\
&Rot. &C-Jit. & Mot.-Blur   & person          & bag         & person         & bag         & person         & bag  \\ 
\hline
%   &SL     & \xmark          & \xmark         & \xmark         &    &      \\
% &SL    & \checkmark         & \checkmark         & \checkmark         &    &      \\
% &SSL    & \xmark         & \xmark         & \xmark         &    &      \\
\multirow{2}{*}{CLASP1}  &\checkmark          & \xmark         & \xmark         &89.2  &43.4  &92.0  &59.7     &83.9   &41.6\\
& \checkmark         & \checkmark         & \checkmark         &91.5   &48.3   &92.3   &64.5  &84.3   &46.4\\
\hline
\multirow{2}{*}{CLASP2}  &\checkmark          & \xmark         & \xmark         &79.4   &47.4 &86.2   &62.5    &73.6   &42.2 \\ & \checkmark         & \checkmark         & \checkmark         &84.2   &47.8     &88.0   &63.0  &76.5   &43.6 \\
\bottomrule
\end{tabular}
\end{table}

% %bag
% \begin{table}[h]
% \centering
% \setlength\tabcolsep{4pt} % default value: 6pt
% %\begin{minipage}{1\textwidth}
% \caption{Ablation for data augmentation in the SSL iteration for bag class.}
% \label{tab:aug_ablation} 
% \begin{tabular}{c|cccccc}
% \toprule
% Dataset  &Rotation & C-Jitter & Motion-Blur &AP & $F_1$ & MODA \\
% \hline
% %   &SL     & \xmark          & \xmark         & \xmark         &    &      \\
% % &SL    & \checkmark         & \checkmark         & \checkmark         &    &      \\
% % &SSL    & \xmark         & \xmark         & \xmark         &    &      \\
% \multirow{2}{*}{CLASP1}  &\checkmark          & \xmark         & \xmark         &43.4   &59.7      &41.6 \\ & \checkmark         & \checkmark         & \checkmark         &48.3    &64.5   &46.4 \\
% %\hline
% \multirow{2}{*}{CLASP2}  &\checkmark          & \xmark         & \xmark         &47.4   &62.5      &42.2  \\ & \checkmark         & \checkmark         & \checkmark         &47.8    &63.0   &43.6 \\
% \bottomrule
% \end{tabular}
% \end{table}
\subsection{Impact of Rotation Resolution}
Table \ref{tab:number_of_rotations_SSL} shows the impact of rotation resolution $r$ on the generation of pseudo-labels. One SSL iteration with $r=20$ improves the MODA scores by up to $3.1\%$ for passengers and $5.6\%$ for baggage items. The inference time for a single frame increases linearly with the number of rotations, contributing to longer SSL training iterations. If training time is a concern, $r=10$ offers a reasonable speed vs. performance trade-off. We use $r=20$ for all the SSL models to demonstrate the potential performance of our framework. As Table \ref{tab:number_of_rotations_SSL} indicates, further increasing the value of $r$ would likely lead to minor additional performance gains.
\begin{table}[h]
\centering
\setlength\tabcolsep{2pt} % default value: 6pt
\caption{Performance impact of the number of rotation angles used in the SSL iterations.}
% Use pre-computed SCT tracklets, no appearance model in MCTA
\label{tab:number_of_rotations_SSL}
\begin{tabular}{c|cccccc}
\toprule
\multicolumn{1}{c}{Dataset} &
\multicolumn{1}{c}{$r$} &
\multicolumn{1}{c}{$\downarrow$Infer. Time} & \multicolumn{2}{c}{$\uparrow$ $F_1$} & \multicolumn{2}{c}{$\uparrow$MODA}\\
 &           &(secs)           & person         & bag         & person         & bag  \\ 
\hline
%   &SL     & \xmark          & \xmark         & \xmark         &    &      \\
% &SL    & \checkmark         & \checkmark         & \checkmark         &    &      \\
% &SSL    & \xmark         & \xmark         & \xmark         &    &      \\
\multirow{4}{*}{CLASP1}  &1  &0.3   &94.5   &69.5  &89.0   &51.7\\
&5  &2.2   &95.0   &70.4  &90.1   &52.8\\
&10  &4.5  &95.3   &70.8  &90.6   &53.6 \\
&20 &9.1   &95.8   &70.8  &91.5   &53.7\\
\hline
\multirow{4}{*}{CLASP2} &1  &0.3   &91.0   &74.5  &82.3  &56.8\\ 
&5  &2.6   &92.1   &76.4  &84.6   &59.6 \\ 
&10  &4.0     &92.1   &76.5  &84.5   &59.8    \\
&20  &11.7   &92.2   &76.5  &84.9   &60.0 \\
\bottomrule
\end{tabular}
\end{table}

\subsection{Semi-Supervised Learning}
\begin{figure}[h]
\centering
\includegraphics[trim=20 20 15 0, clip, height=1.35in]{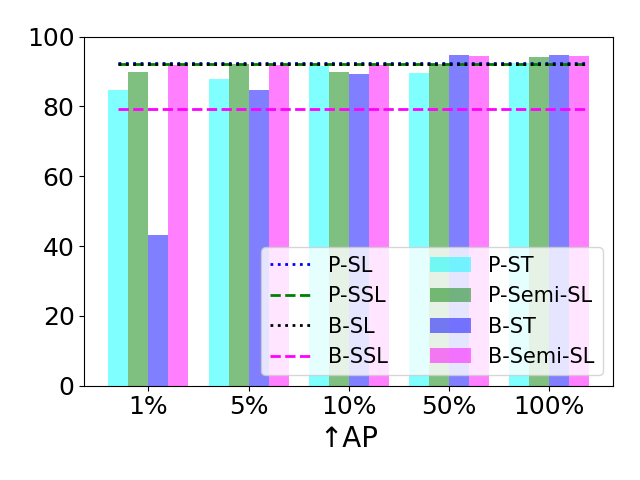}
\includegraphics[trim=60 20 15 0, clip, height=1.35in]{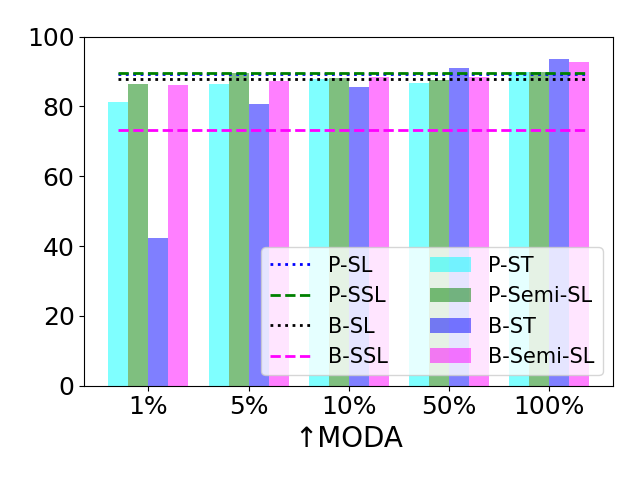}
%Difference between 1 and 5 percent labels in SSL
%\includegraphics[scale=0.94]{baggage_cam11_sota_ours.eps}
%1-iter7
% Average Precision: 0.9177
%  F1  Rcll  Prcn|  FAR     GT     TP     FP     FN| MODA  MODP 
% 93.1  93.1  93.1| 0.00    505    470     35     35| 86.1  79.1
%5-iter7
% Average Precision: 0.9168
%  F1  Rcll  Prcn|  FAR     GT     TP     FP     FN| MODA  MODP 
% 93.7  94.1  93.3| 0.00    505    475     34     30| 87.3  79.2
\caption {Semi-SL model performance on CLASP2 using a semi-supervised extension of our proposed SSL method versus SoftTeacher (ST) \cite{SoftTeacher_2021_iccv}. Here, P and B stand for the passenger and baggage categories. The SSL model uses no labeled data and the SL model is trained with 100\% of the samples.}
\label{fig:semi_learning}
\end{figure}

%% Issue 1: At current iteration, the SSL model might fail to generate the segmentation masks for manual bounding box annotations. We have this issue since we use SSL model from previous iteration to predict the pseudo labels masks where only the bounding boxes are from the manual labels. To tackle this problem, we are ignoring those regressed boxes and manual boxes instead during pseudo-labels predictions [We are hoping the corresponding segmentation masks will be improved iteratively]. 
As Table \ref{tab:detection_person_bag} indicates, the performance of our SSL algorithm is limited by the initial accuracy of the baseline model. Thus, we extend our method to a semi-supervised approach where we use a certain amount of manual annotations to initialize our model before initiating SSL training. For the labeled frames, we employ the same data augmentation procedure used to generate augmented labels. Fig. \ref{fig:semi_learning} shows that training the SSL model using $10\%$ of the manual labels leads to a performance comparable to the SL model, outperforming SoftTeacher \cite{SoftTeacher_2021_iccv}, a state-of-the-art Semi-SL technique. Our method is particularly effective when small amounts of annotations are used. For example, using only $1\%$ of the manual labels, our Semi-SL approach outperforms SoftTeacher by $104\%$ and is only $1.6\%$ behind the SL method (Table \ref{tab:detection_person_bag}) for baggage items. Furthermore, we observe a $5.7\%$ MODA improvement over the SL method when we use all the manual annotations during training.

%%%%%%%%%%%%%%%%%%%%%%%%%%%%%%%%%%%%%%%%%%%%%%%%%%%%%%%%%%%%%%%%
%\begin{comment}
%MOD evaluation measure for person and bag classes
\begin{table*}[h]
\caption{Passenger and baggage detection evaluation measures on the CLASP1 and CLASP2 test sets.}
\label{tab:detection_person_bag}
%\begin{figure}[t]
\centering
\setlength{\tabcolsep}{5pt} % Default value: 6pt
\footnotesize
\begin{tabular}{llcccccccccccccc}
\hline        
\multicolumn{1}{l}{\multirow{1}{*}{Dataset}} & \multicolumn{1}{l}{Model} &
\multicolumn{2}{c}{Method} &
\multicolumn{2}{c}{$\uparrow$Rcll} & \multicolumn{2}{c}{$\uparrow$Prcn} & \multicolumn{2}{c}{$\uparrow$TP} &\multicolumn{2}{c}{$\downarrow$FP} &\multicolumn{2}{c}{$\downarrow$FN} &\multicolumn{2}{c}{$\uparrow$MODA}\\
\multicolumn{2}{c}{}          &$\alpha$ &reg.              & person          & bag         & person         & bag         & person         & bag    & person         & bag     & person         & bag  & person         & bag  \\ 
\hline
\multirow{6}{*}{CLASP1} &Baseline &\xmark &\xmark &73.8 &36.2    &89.0 &87.9    &886 &233    &110 &\bf{32}   &314 &411   &64.7 &31.2 \\ 
                &$\text{SSL}$ &\xmark &\xmark &96.4 &\underline{80.4}    &95.8 &78.6    &1157 &518  &51 &141   &43 &\underline{126}   &92.2 &58.5 \\
                &$\text{SSL}$ &\checkmark &\xmark &\bf{96.9} &70.4   &94.2 &\underline{90.9}   &\bf{1163} &451   &71 &45   &\bf{37} &193  &91.0 &63.0 \\
                &$\text{SSL}$ &\xmark &\checkmark &96.0 &76.1  &\underline{97.7} &90.7  &1152 &490  &\underline{27} &50   &48 &154  &93.8 &68.3 \\
                &$\text{SSL}$ &\checkmark &\checkmark &\underline{96.8} &78.6   &97.3 &90.2   &\underline{1162} &\underline{506}   &32 &55   &\underline{38} &138   &\underline{94.2} &\underline{70.0} \\
                &$\text{SL}$ &\xmark &\xmark &96.7 &\bf{89.4}   &\bf{98.1} &\bf{91.4}  &1160 &\bf{576}   &\bf{23} &54   &40 &\bf{68}  &\bf{94.8} &\bf{81.1} \\
                % \cline{2-16}
                % &$\text{..Semi-ST}$ &\xmark &\xmark &97.2 &84.6   &91.7 &94.5  &1166 &397   &40 &23   &34 &66  &93.8 &80.8 \\
                % &$\text{..Semi-SL}$ &\checkmark &\checkmark &96.1 &88.6   &93.3 &93.6  &1153 &410   &83 &28   &47 &53  &89.2 &82.5 \\

\hline
                       
\multirow{6}{*}{CLASP2}  &Baseline &\xmark &\xmark &73.9 &38.0    &85.1 &78.0     &674 &192  &118 &53    &238 &313   &61.0 &27.5 \\ 
                    &$\text{SSL}$ &\xmark &\xmark &91.2 &67.3    &88.9 &86.5    &832 &340  &104 &53  &80 &165   &79.8 &56.8 \\
                        &$\text{SSL}$ &\checkmark &\xmark &90.1 &\underline{81.6}    &92.9 &81.3    &822 &\underline{412}  &63 &95   &90 &\underline{93}   &83.2 &62.8 \\
                        &$\text{SSL}$ &\xmark &\checkmark &91.3  &70.1    &94.8 &\bf{94.4}    &833 &354     &\bf{46} &\bf{21}   &79 &151    &86.5 &65.9 \\
                        &$\text{SSL}$ &\checkmark &\checkmark &\bf{94.6} &78.6    &\bf{94.8} &93.4    &\bf{863} &397    &\underline{47} &\underline{28}   &\bf{49} &108   &\bf{89.5} &\underline{73.1} \\ 
                    &$\text{SL}$ &\xmark &\xmark &\underline{94.5} &\bf{93.5}    &\underline{94.8} &\underline{94.2}    &\underline{862} &\bf{472}    &47 &29   &\underline{50} &\bf{33}   &\underline{89.4} &\bf{87.7}
                        \\
                % \cline{2-16}
                % &$\text{Semi-ST}$ &\xmark &\xmark &94.8 &87.1   &91.7 &93.2  &865 &440   &78 &32   &47 &65  &86.3 &80.8 \\
                % &$\text{Semi-SL}$ &\checkmark &\checkmark &93.9 &91.9   &95.6 &93.7  &856 &464   &39 &31   &56 &41  &89.6 &85.7 \\

\hline
\end{tabular}
\end{table*}
%\end{comment}

%%%%%%%%%%%%%%%%%%%%%%%%%%%%%%%%%%%%%%%%%%%%%%%%%%%%%%%%%%%%%%%%

%We use camera 2 and 5 frames (different perspectives) for fair comparisons when considering $100\%$ labels in Table. X for training the pre-trained model. 
%Since, our SSL model fine-tunes both the bounding box and segmentation mask heads, we use the fully self-supervised mode to initialize the model about new CLASP1 and CLASP2 datasets and then utilize the manual annotations to update the model.
%TODO: Try similar experiment for COCO datasets used in SoftTeacher
% Bag: CLASP2: 1% labels
% Average Precision: 0.8953, MODA: 86.2

\section{Single-Camera Tracking}
\label{sec:sc-tracking}
\begin{figure*}[h]
\centering
\begin{minipage}{0.9\linewidth}
\centering
%[width=0.5\textwidth] scale = 0.26
\includegraphics[width=0.32\textwidth]{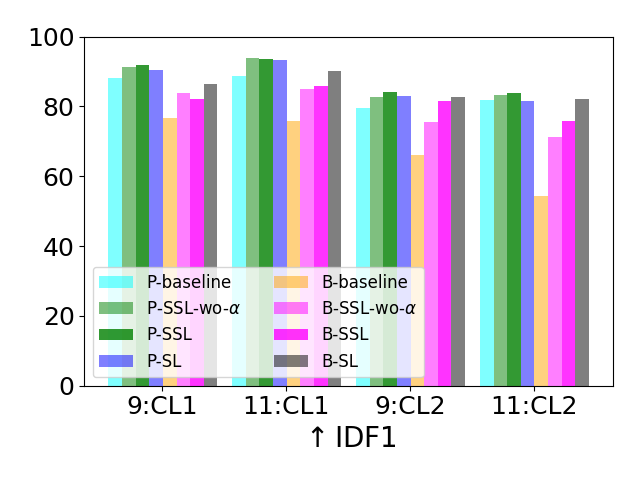}
\includegraphics[width=0.32\textwidth]{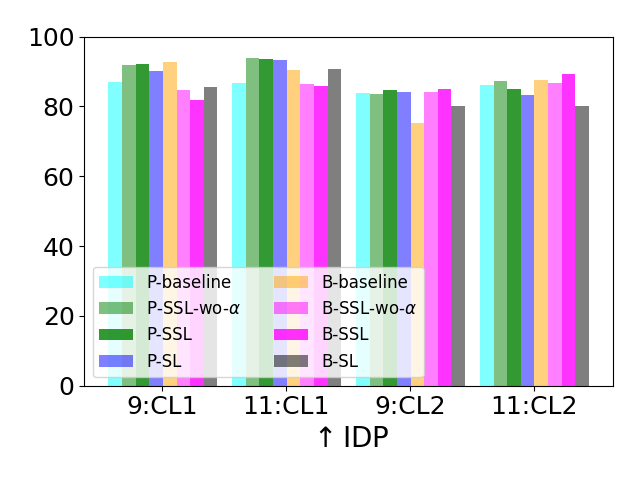}
\includegraphics[width=0.32\textwidth]{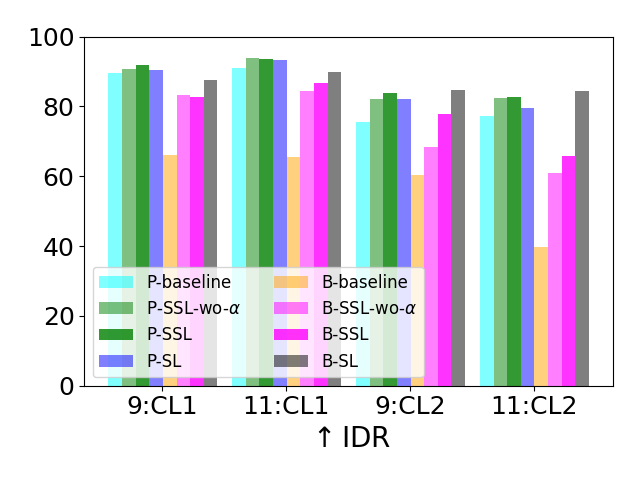} \\
\includegraphics[width=0.32\textwidth]{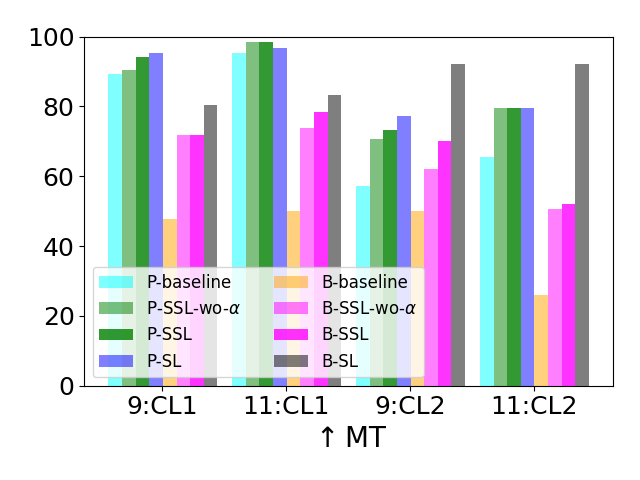}
\includegraphics[width=0.32\textwidth]{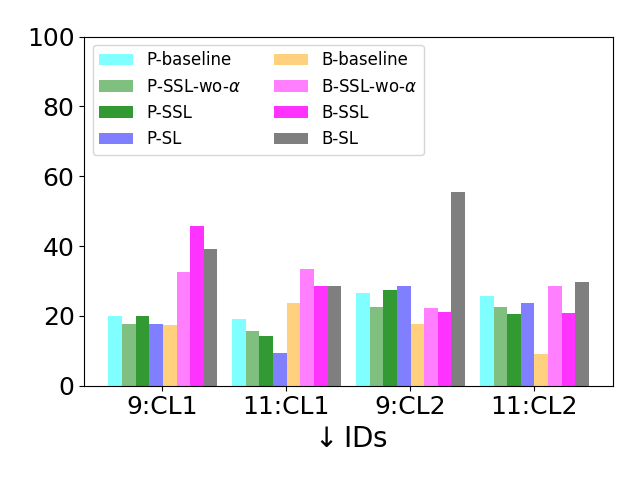}
\includegraphics[width=0.32\textwidth]{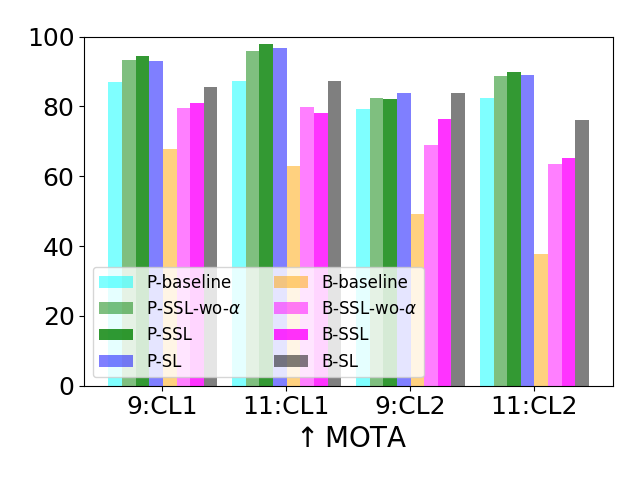} 
\end{minipage}
\caption {Comparison of SCT performance of person and baggage classes in individual cameras of the CLASP1 and CLASP2 datasets using the Baseline, SSL-wo-$\alpha$, SSL, and SL detectors.} %(first four group (left to right) show the performance of MRCNN+MHT  and last four represent the augmented results using our approach).}
\label{fig:track_bar}
\end{figure*}

Fig. \ref{fig:track_bar} shows the Single-Camera Tracking (SCT) performance of our algorithm
%in terms of MOTA, identity based measures \cite{IDF1_scor_eccv}, percentage of mostly tracked trajectories ($\%$MT), 
%fraction of computed detections that are correctly identified (IDR) \cite{IDF1_scor_eccv}, 
%and percentage of id switches ($\%$IDs) 
for passengers and baggage items in the individual cameras of the CLASP1 and CLASP2 datasets. 
For passenger tracking, the SSL methods outperform the SL approach in terms of IDF1, IDP, and IDR in all the scenarios under consideration. In both datasets, the SL approach shows slightly higher MT results for camera 9, largely due to the partial passenger detection problem.
%discussed in Section X of the main paper. 
Since CLASP1 has lower object density, we observe more consistent performance among different methods for both cameras in that dataset. While all the methods perform better on the CLASP1 dataset, the benefits of SSL training compared to the baseline detector are particularly evident in the MT results on the CLASP2 dataset. 

Regarding baggage items, although the SSL models lead to a moderate increase in the number of IDs, these switches are offset by substantial gains in MT. As a matter of fact, the SL model shows a much more significant degradation in IDs for the more complex CLASP2 dataset. This is particularly evident for camera 9, and it explains the lower IDP obtained by the SL method in that dataset. The most evident performance gains for baggage tracking are observed in camera 11 on the CLASP2 dataset because the of the difficulty of partially visible baggage items using the baseline model.

\section{Multi-camera Tracklet Association}
Regarding our Multi-camera Tracklet Association (MCTA) method, Table \ref{tab:ID_Measure_Tracklet_Asso} shows that the Fr\'{e}chet distance metric is particularly useful in crowded scenarios. Although we obtain comparable results using the Hausdorff distance on the easier CLASP1 dataset, we achieve noticeable improvements in all the evaluation criteria on CLASP2 using the Fr\'{e}chet distance. The single-camera trackers in the auxiliary cameras are trained using frames from the primary cameras. Hence, in crowded scenarios they sometimes fails to keep alive trajectories of targets that are temporarily outside the field of view of the primary camera. This is the main reason behind the overall lower tracking performance on the CLASP2 dataset. Training camera-specific detectors using our SSL framework would mitigate this issue.

\begin{table}[h]
\centering
\caption{MCTA evaluation. The column labeled Dist.  indicates whether we employ the Hausdorff ($d_h$) or Fr\'{e}chet ($d_f$) distance to evaluate tracklet similarity.}
\label{tab:ID_Measure_Tracklet_Asso}
\setlength\tabcolsep{1pt}
\begin{tabular}{c|ccccc|ccccc}
\toprule
%we found maximum number of re-entry in camera 2 from 9 (out of scope for evaluation): #re-entry: camera 2 to 9 and camera 5 to 11. Here, #re-entry from camera 9 to 2 and 11 to 5 are not counted to show in evaluation table.
Data. & Dist. &SL &SSL-wo-$\alpha$ &SSL &MCTA  & $\uparrow$IDF1  & $\uparrow$IDR  & $\uparrow$IDP  & $\downarrow$IDs  & $\uparrow$MOTA            \\
 
\hline
\multirow{9}{*}{CLASP1} & -   &\xmark &\checkmark &\xmark  &\xmark &87.4  &87.8  &86.9  &45  &94.4    \\
                    & -    &\xmark &\xmark &\checkmark  &\xmark &87.0  &87.4  &86.6  &48  &95.1   \\
                    & -    &\checkmark &\xmark &\xmark  &\xmark &87.3  &87.5  &87.2  &42  &94.4    \\
                    
                    \cline{2-11} 
                    &  \multirow{3}{*}{$d_h$}
                     &\xmark &\checkmark &\xmark &\checkmark  &\bf{94.0}  &\bf{94.5}  &\bf{93.5}  &\bf{21}  &94.9    \\
                    &    &\xmark &\xmark &\checkmark &\checkmark  &92.7  &93.1  &92.2  &30  &\underline{95.5}    \\ 
                &  &\checkmark &\xmark &\xmark &\checkmark  &93.0  &93.1  &92.8  &25  &94.8    \\

 \cline{2-11}
 
                   & \multirow{3}{*}{$d_f$}
                    &\xmark &\checkmark &\xmark &\checkmark  &\underline{93.8}  &\underline{94.3}  &\underline{93.3}  &\underline{22}  &94.9   \\
                   
                   &   &\xmark &\xmark &\checkmark &\checkmark  &92.8  &93.3  &92.4  &27  &\bf{95.6}    \\
                   &  &\checkmark &\xmark &\xmark &\checkmark  &93.4  &93.6  &93.3  &23  &94.8    \\
\cline{1-11}
\multirow{9}{*}{CLASP2}    & -     &\xmark &\checkmark &\xmark &\xmark     &76.9  &78.7  &75.1  &112  &82.0  \\
                        & -     &\xmark &\xmark &\checkmark &\xmark     &77.0  &78.8  &75.3  &122  &81.9 \\
                        & -     &\checkmark &\xmark &\xmark &\xmark     &75.8  &78.2  &73.6  &128  &81.6  \\
                        \cline{2-11} 
                        
                    &  \multirow{3}{*}{$d_h$}  &\xmark &\checkmark &\xmark &\checkmark  &81.2  &83.3  &79.3  &\underline{94}  &82.8    \\
                    &   &\xmark &\xmark &\checkmark &\checkmark  &82.1  &84.2  &80.3  &110  &82.5    \\
                    &  &\checkmark &\xmark &\xmark &\checkmark  &79.5  &82.0  &77.2  &109  &82.5    \\

\cline{2-11}                    
                & \multirow{3}{*}{$d_f$}   &\xmark &\checkmark &\xmark &\checkmark  &\underline{82.3}  &\underline{84.4}  &\underline{80.4}  &\bf{93}  &\underline{83.0}    \\
                &   &\xmark &\xmark &\checkmark &\checkmark  &\bf{83.6}  &\bf{85.7}  &\bf{81.7}  &105  &82.7    \\
                &   &\checkmark &\xmark &\xmark &\checkmark  &80.1  &82.7  &77.8  &99  &\bf{83.2}    \\
\bottomrule
\end{tabular}
\end{table}

% \section{Conclusion}
% \label{sec:conclusion}
% In this work, 

%Our event detection algorithm record most of the events related to the passengers activities in the security checkpoints.
%TODO: We may want to remove this before submission
% To generate the table, run:
% pdflatex n.tex
% makeindex egpaper_for_review.nlo -s nomencl.ist -o egpaper_for_review.nls
% pdflatex n.tex
%\printnomenclature

\section{Computational Complexity}
In this section, we analyze the theoretical computational complexity of our SSL strategy and measure the computation time and memory utilization of each step of our algorithm. All our experiments were performed on a workstation equipped with two RTX-2090Ti GPUs and an Intel$^{\circledR}$
 Xeon$^{\circledR}$
 Silver 4112 CPU $@ 2.6G$Hz. 

\subsection{Self-Supervised Learning}
The computational complexity of our approach increases linearly with the number of rotation angles used for augmentation in the pseudo-label generation step. That is, for a baseline detection algorithm with computational complexity $\Theta(f(I(t))$, the complexity of our approach is $\Theta(r\cdot f(I(t))$, where $r$ is the number of rotation angles. For example, for $r=20$, the run-time is 20 times that of a single iteration without augmentation. However, these operations are parallelizable as long as the hardware resources support the simultaneous processing of multiple frames.  With our unoptimized implementation, the total time to complete one SSL iteration is approximately six hours for both model training and pseudo-label generation. However, we have observed that hardware resources are severely underutilized, which indicates substantial room for reduction in overall computation time.

\subsection{Inference Performance}
Table \ref{tab:all_complexity_summary} shows the computation time of the proposed tracking-by-detection algorithm, employing a PANet detector with a ResNet-50 backbone. The SCT uses the detector results and a ResNet-50-based Re-Identification (Re-ID) model trained on MOT17 to re-label tracklets lost due to short-term occlusions. Hence, the computation time and memory utilization for the SCT are similar to those for the detector model. Since we are processing single images individually instead of image batches, the inference time for the detector and the SCT are far from optimal. Preliminary experiments indicate that processing batches of 10 images simultaneously leads to an approximate six-fold reduction in detector inference time without exceeding the memory capacity of the GPUs. Reusing the backbone features from the detector in the Re-ID model should also lead to a dramatic reduction in SCT time, since feature generation is the most computationally demanding element of the tracking algorithm.

%complexity: full algorithm, offline-MCTA
\begin{table}[h]
\centering
\caption{Computation time of the proposed tracking-by-detection framework.}
% MCTA: Freachet; max track size: --
\label{tab:all_complexity_summary}
\setlength\tabcolsep{3pt}
\begin{tabular}{c|ccc}
\toprule
Data &Model    & Infer. Time (ms) & Memory (MB) \\
\hline
\multirow{3}{*}{CLASP1} &Detector & 333.3   &1,850         \\
&SCT      & 142.8                 &1,748          \\
&MCTA     & 25.6                &9.1      \\
\hline
\multirow{3}{*}{CLASP2} &Detector & 333.3       &1,850         \\
&SCT      & 166.6               &1,750          \\
&MCTA     & 83.3                &22.7     \\
\bottomrule
\end{tabular}
\end{table}
%Hausdorff
%clasp1: no track size
% >> Primary Frames:  5048.8
% >> Auxiliary Frames:  4104.6
% >> Primary Tracks:  57.0
% >> Auxiliary Tracks:  47.6
% >> Inference Time FPS:  1923.0
% >> CPU Memory:  4535243.4
%tracksize 240
% >> Primary Frames:  5048.8
% >> Auxiliary Frames:  4104.6
% >> Primary Tracks:  57.0
% >> Auxiliary Tracks:  47.6
% >> Inference Time FPS:  1967.0
% >> CPU Memory:  4554316.8

%clasp2: no track size
% >> Primary Frames:  11335.333333333334
% >> Auxiliary Frames:  10838.666666666666
% >> Primary Tracks:  158.33333333333334
% >> Auxiliary Tracks:  140.0
% >> Inference Time FPS:  1551.6666666666667
% >> CPU Memory:  16310136.333333334
%track size <=240
% >> Primary Frames:  11335.333333333334
% >> Auxiliary Frames:  10838.666666666666
% >> Primary Tracks:  158.33333333333334
% >> Auxiliary Tracks:  140.0
% >> Inference Time FPS:  1631.0
% >> CPU Memory:  16310136.333333334

The execution time of the proposed MCTA algorithm depends on the average length of the overlapping tracklet segments in each camera pair. In the CLASP1 dataset, which contains fewer and shorter tracklets, the algorithm can be executed in real time. In CLASP2, it can run at approximately 12 fps. However, the current implementation of the proposed system uses the full life-span of a tracklet to compute the Fr\'{e}chet association distance in the MCTA algorithm. It is possible to substantially reduce computation time by limiting the length of single-camera tracklets compared by the algorithm. Table \ref{tab:mcta_complexity_summary} shows that if we limit the length of the tracklets to $240$ frames (or eight seconds), it is possible to achieve real-time performance for both datasets without degrading the accuracy of the algorithm. 

%For example, we measure the computational complexity of MCTA for tracklet matching between camera 9 and camera 2 in CLASP1 and CLASP2 datasets, where the average number of tracklets for Hungarian matching is $[62, 81]$, $[140, 158]$ and the intermediate frames where the tracklets exist are $[4870, 6341]$, $[10838, 11335]$ respectively. Here we use the SSL detector with $\alpha$ and $\text{reg.}$ method to get the single-camera tracklets. We observed that the maximum $8$ seconds life-span for a track from $30$ FPS input videos boost the MCTA speed by a maximum $200\%$ in terms of frames per second and the CPU memory utilization by $39.3\%$ without degrading the overall tracking metric MOTA.
%complexity: offline-MCTA
\begin{table}[h]
\centering
\caption{Computation time of the proposed MCTA.}
% Use pre-computed SCT tracklets, no appearance model in MCTA
\label{tab:mcta_complexity_summary}
\setlength\tabcolsep{2pt}
\begin{tabular}{c|cccccc}
\toprule
Data. &Dist. Metric             & Max. Size & Infer. Time (ms) & Memory (MB) &MOTA\\
\hline
\multirow{4}{*}{CLASP1} &\multirow{2}{*}{Hausdorff}  & --    &0.52   &4.5   &94.5 \\
                        &    &240   &0.50   &4.5   &94.5 \\
&\multirow{2}{*}{Fr\'{e}chet} & --   &25.6     &9.1   &94.6 \\
                    &     &240   &8.20      &4.3   &94.6 \\
\hline
\multirow{4}{*}{CLASP2} &\multirow{2}{*}{Hausdorff}  & --    &0.64   &16.3   &78.4 \\
                   &         & 240   &0.61  &16.3   &78.4 \\
&\multirow{2}{*}{Fr\'{e}chet} & --   &83.3     &22.7   &78.5 \\
                     &    & 240   &19.6       &16.3   &78.7 \\
\bottomrule
\end{tabular}
\end{table}

\end{document}